\documentclass[lettersize,journal]{IEEEtran}
\usepackage{amsmath,amsfonts}
\usepackage{algorithmic}
\usepackage{algorithm}
\usepackage{array}
\usepackage[caption=false,font=normalsize,labelfont=sf,textfont=sf]{subfig}
\usepackage{textcomp}
\usepackage{stfloats}
\usepackage{url}
\usepackage{verbatim}
\usepackage{graphicx}
\usepackage{cite}

\usepackage{color, colortbl}

\begin{document}

\title{Impact of Strategic Sampling and Supervision Policies on Semi-supervised Learning}

\author{Shuvendu~Roy,~\IEEEmembership{Student Member,~IEEE,}
        and~Ali~Etemad,~\IEEEmembership{Senior Member,~IEEE}
\IEEEcompsocitemizethanks{\IEEEcompsocthanksitem S. Roy and A. Etemad is with the Department of ECE and Ingenuity Labs Research Institute, Queen's University, Canada.
\protect\\
E-mail: \{shuvendu.roy; ali.etemad\}@queensu.ca
}
}
 
\maketitle

\begin{abstract}
In semi-supervised representation learning frameworks, when the number of labelled data is very scarce, the quality and representativeness of these samples become increasingly important. Existing literature on semi-supervised learning randomly sample a limited number of data points for labelling. All these labelled samples are then used along with the unlabelled data throughout the training process. In this work, we ask two important questions in this context: (1) does it matter which samples are selected for labelling? (2) does it matter how the labelled samples are used throughout the training process along with the unlabelled data? To answer the first question, we explore a number of unsupervised methods for selecting specific subsets of data to label (without prior knowledge of their labels), with the goal of maximizing representativeness w.r.t. the unlabelled set. Then, for our second line of inquiry, we define a variety of different label injection strategies in the training process. Extensive experiments on four popular datasets, CIFAR-10, CIFAR-100, SVHN, and STL-10, show that unsupervised selection of samples that are more representative of the entire data improves performance by up to $\sim$2\% over the existing semi-supervised frameworks such as MixMatch, ReMixMatch, FixMatch and others with random sample labelling. We show that this boost could even increase to 7.5\% for very few-labelled scenarios. However, our study shows that gradually injecting the labels throughout the training procedure does not impact the performance considerably versus when all the existing labels are used throughout the entire training.
\end{abstract}


\section{Introduction}
The availability of large-scale annotated datasets has significantly contributed to the success of deep learning \cite{krizhevsky2012imagenet, vaswani2017attention}. However, data annotation and labelling are often time-consuming and expensive. As a result, the benefits of supervised deep learning come at the cost of collecting large-scale labelled datasets.  It is, therefore, long desired to train deep learning models without using such large labelled datasets. As one of the potential solutions to this problem, semi-supervised learning (SSL) \cite{fixmatch,mixmatch} has gained increased popularity in recent years. Semi-supervised methods learn from small labelled data while exploiting large amounts of unlabelled data (which is often readily available) to improve the overall performance.

Generally, semi-supervised methods randomly select a few samples from a large set of unlabelled data and label/annotate those samples to create a small labelled set \cite{fixmatch,mixmatch,ssl_fer}. However, these methods don't consider the fact that all samples are not equally representative of the underlying data distribution. With a limited labelling budget, selecting samples to label can influence the model's final performance; thus, strategic labelling is crucial.
Another common practice in existing semi-supervised methods is to use all the labelled samples uniformly throughout the training process. 
To date, no prior work has studied the strategy with which to inject the small amounts of labelled data along with the large unlabelled sets. These two problems motivate us to ask the following two questions: \textbf{(1)} 
Can we select a specific set of unlabelled samples for labelling to enhance the performance of the semi-supervised learner compared to random sampling?
\textbf{(2)} Is there a more effective strategy to incorporate labelled data into the semi-supervised training pipeline?

In this work, we investigate the two problems mentioned above. First, we explore several unsupervised methods for strategically selecting key samples from the pool of unlabelled data to label. Although subset selection is not a new concept in the literature \cite{dist_density_tails, data_diet}, (\textit{a}) they have not been explored in the context of semi-supervised learning, and (\textit{b}) labels are often utilized for finding these subsets, effectively making them supervised selection methods; whereas our interest lies in unsupervised selection given that we start with a large set of unlabelled data. 
\textcolor{black}{
Given a labelling budget of `$n$' samples, we first extract the embeddings of all the unlabelled data using a pre-trained encoder and then partition the embedding space into $n$ clusters. Once the clusters are formed, the closest sample to the centroid from each cluster is selected for labelling. Since clustering divides the embedding space into distinct clusters based on their inherent similarities and dissimilarities, selecting a sample per cluster will ensure the overall diversity of the unlabelled dataset. 
The strategic sampling process is illustrated in Figure \ref{fig:cluters}. 
Here, we show an example of selecting 6 samples from the unlabelled set. First, the embedding space is partitioned into 6 clusters (red dots represent the centroids), and the closest sample to each centroid is then selected to be labelled.
}

Once a few samples are selected and labelled, we investigate our second question, i.e, whether using all the labelled samples together from the beginning of the training process is the best approach or whether a well-defined injection strategy can improve the performance. Accordingly, we explore a number of supervision policies that utilize the labelled samples at different points throughout the training stage. These strategies include consistent increments of labelled data and also curriculum-based label injection strategies.

Given a fixed amount of labelling budgets, we show that the accuracy of any semi-supervised method can be improved by up to $\sim$2\% by replacing the randomly sampled labelled set with the proposed label set selection strategy. Similar experiments in supervised settings also show that the proposed strategic sampling outperforms randomly selected labelled sets in fully supervised training. We demonstrate that strategic sampling can be used as an easy-to-add extension to any off-the-shelf semi-supervised method to improve performance. Also, note that the addition of this approach does not add any computational overhead for the semi-supervised training protocol. It only introduces a one-time computation at the beginning of training. For our second research question, we find that the incremental/curriculum supervision policies do not considerably improve performance. In summary, we make the following contributions in this paper. 
\begin{itemize}
\item 
We demonstrate that the unsupervised selection of the most representative samples to label and use alongside the unlabelled set can improve semi-supervised learning. 
\item 
The proposed strategic sampling can be easily integrated into any semi-supervised learner. Experiments show improvements over existing semi-supervised methods with the introduction of our proposed method without any increase in model size or FLOPS.
\item
We investigate different supervision policies to systematically introduce the labelled samples during semi-supervised training. However, our findings suggest no considerable improvements over the current approach of using labelled samples, i.e., using all the labelled samples throughout the entire training.
\end{itemize}

\section{Related Work}\label{sec:related_work}
In this section, we summarize prior works in two key areas relevant to our problem statements: semi-supervised learning and data sampling.

\subsection{Semi-Supervised Learning}
\textcolor{black}{
Unsupervised learning is a branch of machine learning in which algorithms extract the inherent patterns and structures within the dataset without any supervision \cite{simclr,fixmatch}. Examples of unsupervised learning methods are clustering, dimensionality reduction, self-supervised learning, and others. Recently, there has been increased interest in unsupervised learning, mainly due to the strong performances obtained via self-supervised \cite{simclr} and semi-supervised frameworks \cite{fixmatch,remixmatch}. Self-supervised methods learn from the unlabelled data by defining an auxiliary task, allowing models to extract meaningful representations from the unlabelled data. The learned representation can be adapted to a downstream task with minimal supervision. While self-supervised methods learn from the unlabelled data alone, semi-supervised learning assumes a small amount of labelled data is also available during training, requiring both supervised and unsupervised learning in the same framework.}

Most successful methods for semi-supervised learning can be categorized into two broad categories: consistency regularization \cite{pi_model,vat,mean_teacher} and entropy minimization \cite{grandvalet2004semi,pseudo_labels}. The consistency regularization methods are based on the concept of \textit{data transformations}, with the assumption that the underlying class semantic information does not change when different transformations are applied. Under two augmentations applied to a sample, a model should generate the same class prediction. More formally, in its simplest form \cite{pi_model,laine2016temporal}, consistency regularization can be presented as:
    \begin{equation}
        ||P_\theta(y~|~\text{Aug}_1(x) - P_\theta(y~|~\text{Aug}_2(x)||_2^2, 
    \end{equation}
where $P_\theta$ is the prediction of the model on unlabelled data $x$ when augmentation $Aug$ is applied. Consistency regularization-based methods also use a supervised loss on the small amounts of labelled data to learn the class predictions. 

Pi-model \cite{pi_model} is one of the popular methods in this category. It forces the representations of two augmented images to be similar by applying consistency regularization between them. Later, MeanTeacher \cite{mean_teacher} improved this method by utilizing an exponential moving average (EMA) on the encoder model. Here, the learned encoder is called the student encoder, and the EMA of the student is called the teacher encoder. In MeanTeacher, one of the two augmented images is encoded by the student encoder, whereas the other image is encoded by the teacher. This approach learns by regularizing the consistency between these two embeddings. The loss function for this method is as follows:  
\begin{equation}
        ||P_\theta(y~|~\text{Aug}_1(x) - P_{EMA(\theta)}(y~|~\text{Aug}_2(x)||_2^2. 
    \end{equation}
Two other improved variants of the consistency regularization method are VAT \cite{vat} and UDA \cite{uda}. In VAT, the concept of adversarial perturbation was used as an alternative to augmentation. Therefore, the resulting methods regularized the consistency between an image and its adversarial perturbation. UDA utilized the concept of hard augmentation instead of simpler augmentations and showed that semi-supervised learning benefits from hard augmentations. More specifically, UDA regularized between an input image and a hard augmentation of that image.
\textcolor{black}{Adaptive Neighborhood Propagation \cite{jia2016adaptive} proposed a different approach that integrates sparse coding and label propagation, avoiding the need to manually choose neighbourhood size or kernel width that some of the previous methods used for pseudo-labelling.}

Entropy minimization methods generally aim to generate low entropy predictions.
This is done by either explicitly defining a loss on the entropy \cite{grandvalet2004semi} or implicitly by generating predictions in the form of pseudo-labels \cite{pseudo_labels}. This is also combined with a supervised loss on the labelled data. One example of this category of methods is the Pseudo-label technique \cite{pseudo_labels}. This approach learns from the labelled data in a supervised setting and predicts the pseudo-labels for the unlabelled samples. If the confidence of the pseudo-labels is beyond a pre-defined threshold, these images and predicted pseudo-labels are added to the labelled set, effectively increasing the size of the labelled set.

Recently, some of the most successful methods combined consistency regularization with entropy minimization in a hybrid framework \cite{mixmatch,remixmatch,fixmatch}, which showed improved performances. An example is MixMatch \cite{mixmatch}, where the loss function consists of an entropy minimization term and a consistency regularization term. This method popularized the use of the MixUp operation, which interpolates between labelled and unlabelled images from inputs and labels. ReMixMatch \cite{mixmatch} is an improvement on MixMatch with the introduction of two new concepts: distribution alignment and augmentation anchoring. The distribution alignment forces the predicted marginal distributions of unlabelled data to be close to that of the labelled data. Meanwhile, the augmentation anchoring forces the representation of multiple augmentations to be consistent with that of a weakly augmented image. Later, FixMatch \cite{fixmatch} proposed a simple idea that showed very good results across different datasets and training setups. This method used a weakly augmented image to predict the pseudo-label for an input image. The pseudo-label is retained only if the confidence is higher than a pre-defined threshold. The model is then trained to predict the pseudo-label for a hard augmentation of the same image. 
More recently, FlexMatch \cite{zhang2021flexmatch} introduced the concept of curriculum to adjust the threshold for selecting pseudo-labels. SoftMatch \cite{chen2023softmatch} proposed to maintain a balance between the quantity and quality of pseudo-labels during training, using a truncated Gaussian function to weight samples based on their confidence. FullMatch \cite{fullmatch} improved the unlabelled data usage by proposing two novel techniques, Entropy Meaning Loss (EML) and Adaptive Negative Learning (ANL), where EML generates high-confidence predictions, and ANL adaptively allocates negative pseudo-labels to low-confidence examples. Finally, UnMixMatch \cite{roy2024scaling} improved the scalability of semi-supervised learning by learning from unconstrained unlabeled data.
Some of the methods also introduced improvements to the model and training pipelines. For example, \cite{zhang2021dual} proposed a multi-layer, hierarchical representation learning framework incorporating label and structure constraints for improved feature representations. AELP-WL \cite{zhang2017robust} seamlessly combines adaptive embedded label propagation with adaptive weight learning in a unified framework to jointly optimize weights for representation learning and classification. 

In both consistency regularization and entropy minimization methods, a small labelled set plays a critical role in guiding the learning using unlabelled data. As a result, the quality and representativeness of the labelled data are very important. To the best of our knowledge, no prior method has attempted to select the most representative and diverse samples from an unlabelled set of samples for semi-supervised training. 

\subsection{Data Sampling}
Prior works focusing on data sampling often aim at selecting a small but useful subset from a large dataset \cite{coleman2019selection,data_diet}. In other words, these methods aim to approximate the distribution of the whole dataset using only a small subset of the data. Most of these methods rely on the `labels' to find the most representative subsets, effectively making them `supervised' methods. These methods start with the labelled dataset and find useful samples early in the training to reduce the training cost with the whole dataset. For instance, in \cite{coleman2019selection}, samples with the least confidence in class prediction were preferred with the intuition that the least certain samples gave a greater impact on the optimization. Some works 
use the loss or errors as a signal for selecting subsets. For instance, \cite{Forgetting} counts the number of times a model forgets the prediction for a sample (misclassifies a sample that was previously classified correctly). The selection strategy then removed the unforgettable samples to form the subset. In \cite{data_diet}, the contribution of each sample towards the decline of the loss was calculated over several individual runs to select the intended subset. 

Another popular concept for subset selection is an incremental selection which starts with a small subset of random samples. This is also known as active learning. For example, in \cite{margatina2021active}, the samples with the highest likelihood of divergence
w.r.t. their neighbours are selected. Another popular approach is to select samples that are close to decision boundaries. Since the actual distance of a sample from the decision boundary can not be measured, an adversarial perturbation-based approach was proposed to approximate this in \cite{ducoffe2018adversarial}. This method applied perturbations to a sample and counted the number of perturbations needed for the class prediction to change. Those with the smallest counts were identified to be the closest to the decision boundary. Active learning is also supervised as the labels of all the samples are used in the selection process. In a semi-supervised learning setting, we need to select a subset from the `unlabelled' set for labelling. Consequently, an `unsupervised' subset selection method needs to be used to find the desired subset from the `unlabelled' samples.

\begin{figure}
    \centering
    \includegraphics[width=0.5\textwidth]{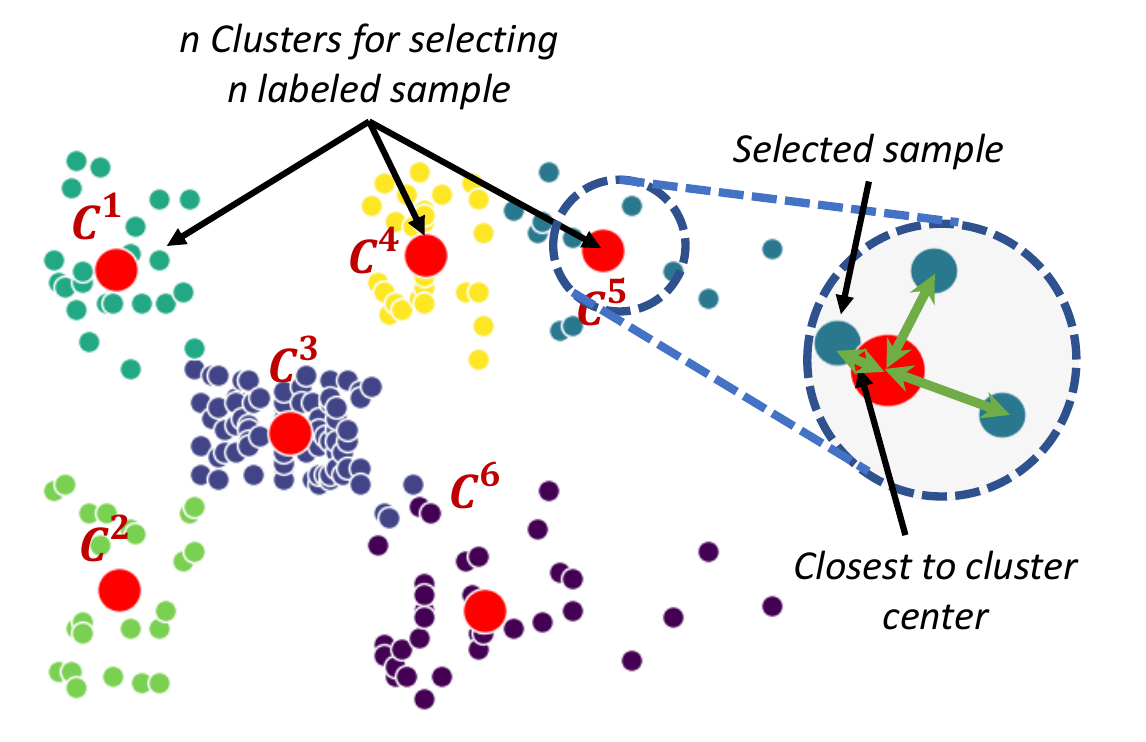}
    \caption{\textcolor{black}{Visual illustration of strategic sampling for a labelling budget of 6 samples. We first cluster the embedding space into 6 clusters (red dots represent the cluster centres) and then select the sample closest to each centroid for labelling.}}
    \label{fig:cluters}
\end{figure}

\section{Method}\label{sec:method}
This section first describes the preliminaries on semi-supervised learning and the problem statement, followed by descriptions of strategic sampling and incremental supervision policy, which we aim to inquire in this paper. 

\subsection{Preliminaries}
Let, $X_U={(x_i)_{i=1}^N}$ be a set of unlabelled samples and $X_L={(x_i,y_i)_{i=1}^n}$ be a subset of $X_U$ that contains labels as well. In the semi-supervised setting, we assume $n \ll N$. The number of classes in the dataset is denoted by $c$. 
Semi-supervised learning aims to utilize $X_U$ using an unsupervised loss, while learning from $X_L$ in a supervised setting. 
Formally, semi-supervised learning can be represented with the following equation:
\begin{equation}
		\min_{\theta} \underset{\text{supervised loss}}{\underbrace{\sum_{(x,y)\in X_L}\mathcal{L}_S(x,y,\theta)}}+\alpha  \underset{\text{unsupervised loss}}{\underbrace{\sum_{x\in X_U}\mathcal{L}_U(x,\theta)}}~,
		\label{equ: semiLoss}
	\end{equation}
where $\theta$ is the model parameters, $L_S$ is the supervised loss, and $L_U$ is the unsupervised loss. The unsupervised loss is the most unique and distinguishable component across different semi-supervised methods.

\subsection{Problem Formulation}
Given, the unlabelled set $X_U$, existing semi-supervised learning literature samples a random set of data points to create the labelled set $X_L$. However, in the unlabelled set $X_U$, not every sample is well representative of the underlying class distribution. Also randomly selecting a set of samples does not ensure a diverse set. As a result, we hypothesize that randomly selecting the $n$ samples to label may not necessarily yield the best performance for the model. However, it is generally challenging to sample an effective subset without any knowledge of the actual labels. This forms the basis of our first line of inquiry.

As for our second research question, we notice that a common practice in semi-supervised learning is to utilize all the labelled data throughout the entire training process. As a result, we aim to explore whether this is indeed the most effective approach for injecting the labelled data throughout semi-supervised representation learning.

Following, we provide formal definitions for the two questions which we explore in this paper:

\subsubsection{Definition 1: strategic sampling} 
Given the unlabelled dataset $X_U={(x_i)_{i=1}^N}$, we aim to sample a subset $X_l={(x_i)_{i=1}^n}$, $1 \leq n < N$, to form the labelled set $X_L={(x_i,y_i)_{i=1}^n}$ for downstream use by the SSL method $\mathcal{S}$ along with the unlabelled set $X_U$. The goal of \textit{strategic sampling} is to select the most diverse and representative set of samples $X^{*}_L$ for labelling, which can provide a performance boost to the semi-supervised learner. 

\subsubsection{Definition 2: Supervision Policy} 
To train an SSL backbone $\mathcal{S}$, the labelled subset $X^*_L$ can be used according to a supervision policy $P$, where $P$ dictates: (\textit{a}) how many labelled samples $X^P_L={(x_i,y_i)_{i=1}^p} \in X^*_L$ are used at epoch $e$, and (\textit{b}) the order in which $X^P_L$ are used in the training process.

\subsection{Explored Solutions}

\subsubsection{Strategic Sampling Module}
Given the labelling budget of $n$ samples, the strategic sampling module aims to select the samples that provide the maximum final performance for the SSL method. For selecting $n$ samples from the unlabelled set $X_U$, the strategic sampling module utilizes a pre-trained encoder $\theta_p$, which extracts the embedding for each of the samples in $X_U$. Using pre-trained models for feature extraction is a well-established practice in the literature \cite{zhuang2020comprehensive}.
In addition to exploring the final embeddings produced by the encoder, we also use the intermediate representations throughout our experiments. For each $x_i\in X_U$, the embedding $\hat{z_i}$ is generated as:
    \begin{equation}
        \hat{z_i} = P(x_i, \theta_p).
    \end{equation}

Given the generated embeddings ${(\hat{z_i})_{i=1}^N}$, we utilize a clustering method to partition the embeddings into `n' distinctive clusters. Here, $n$ is the number of samples (annotation budget) we aim to select. The clustering method returns ${(C^i)_{i=1}^n}$ representing $n$ cluster centroids. Once the clusters are formed, a data point is selected from each of the clusters (total $n$ samples) to create the selected labelled set $X^{*}_L$. We use the Euclidean distance to select the nearest sample to each of the cluster centers. A visual illustration of the strategic sampling process is shown in Figure \ref{fig:cluters}. Since clusters are formed by positioning similar samples together, selecting one sample (near the cluster center) from each cluster results in a diverse set of samples to train the model with.

In this work, we use K-means and Bisecting K-means \cite{steinbach2000comparison} as the clustering algorithms. We also utilize the well-initialized versions known as K-means++ and Bisecting K-means++ to form better clusters. The K-means algorithm is one of the most popular and widely used clustering algorithms in the literature. In K-means, we first define the number of clusters. Each data point is then assigned to one of the clusters based on the distance from the cluster centers (centroid). The objective is to assign all the data points to the nearest centroid while keeping the clusters as small as possible. The term `means' in the K-means algorithm comes from the fact that the centroids are calculated as the average of the data points currently belonging to this cluster. This simple clustering technique is shown to be useful for many different domains. However, naive K-means has some limitations, which are addressed by different improvements, for example, K-means++.

In naive K-means, the initial location of the centroids can dictate the outcome of the clustering algorithm. It has been shown that a bad initialization can cause the model to find very different centroids than one with better initialization \cite{arthur2006k}. To solve this issue, the K-means++ algorithm proposed an effective initialization technique for cluster centers while keeping the rest of the algorithm identical to the original K-means \cite{arthur2006k}. K-means++ selects the first centroid randomly from all the data points. It then calculates the distance of all the data points from the already assigned centroid (or centroids). The data point with the highest distance from all the existing centroids is considered the next centroid. This step continues until the desired number of centroids is found. The bisecting K-means clustering technique is another small modification of the original K-means algorithm \cite{savaresi2001performance}. In this algorithm, all points are first divided into two clusters (hence the name bisecting). Then each cluster is further divided into two more clusters until the desired number of centroids is achieved. This algorithm also has a better initialization version called bisecting K-means++.

\textcolor{black}{Applying K-means on an embedding space has been considered in prior works for dimensionality reduction of the feature space \cite{boutsidis2014randomized,cohen2015dimensionality}. This indicates that the K-means family of clustering methods can effectively reduce the redundancies in the embedding space. Moreover, our visual analysis (presented later in Section \ref{sec:res:clustering}), shows that indeed selecting the centroids of clusters as representative items is a viable solution given the evident similarities between data points located near the centroids in the embedding space.}

\begin{figure}
\centering
\subfloat[]{\includegraphics[width= 0.33\columnwidth]{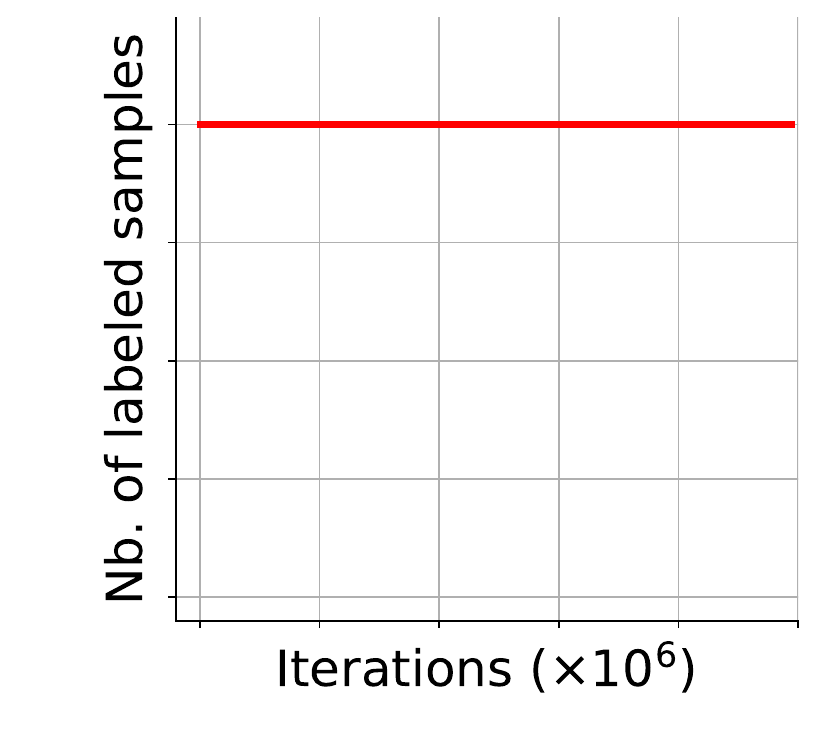}}
\subfloat[]{\includegraphics[width= 0.33\columnwidth]{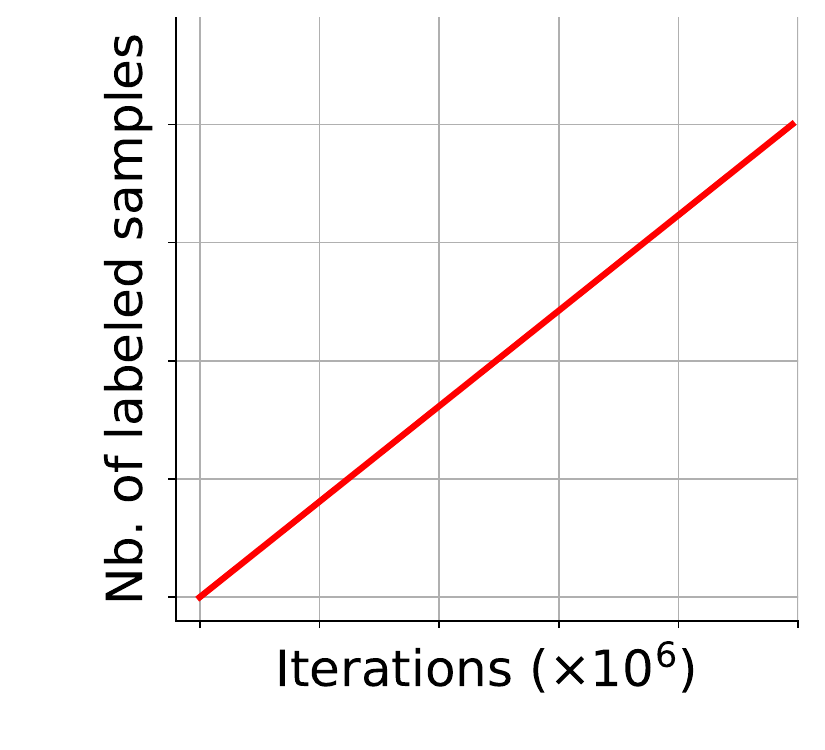}}
\subfloat[]{\includegraphics[width= 0.33\columnwidth]{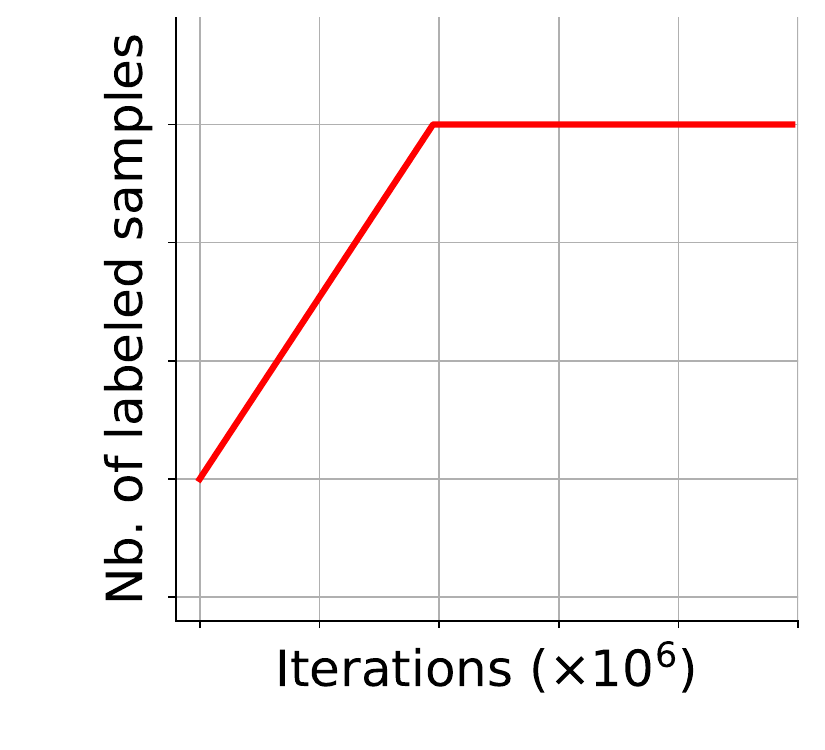}}\\
\subfloat[]{\includegraphics[width= 0.33\columnwidth]{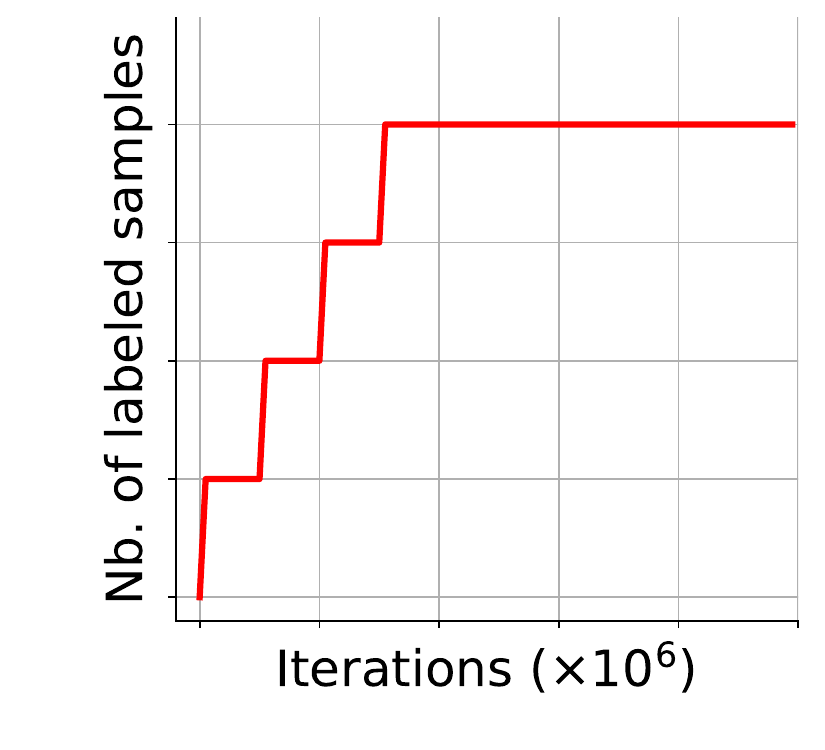}}
\subfloat[]{\includegraphics[width= 0.33\columnwidth]{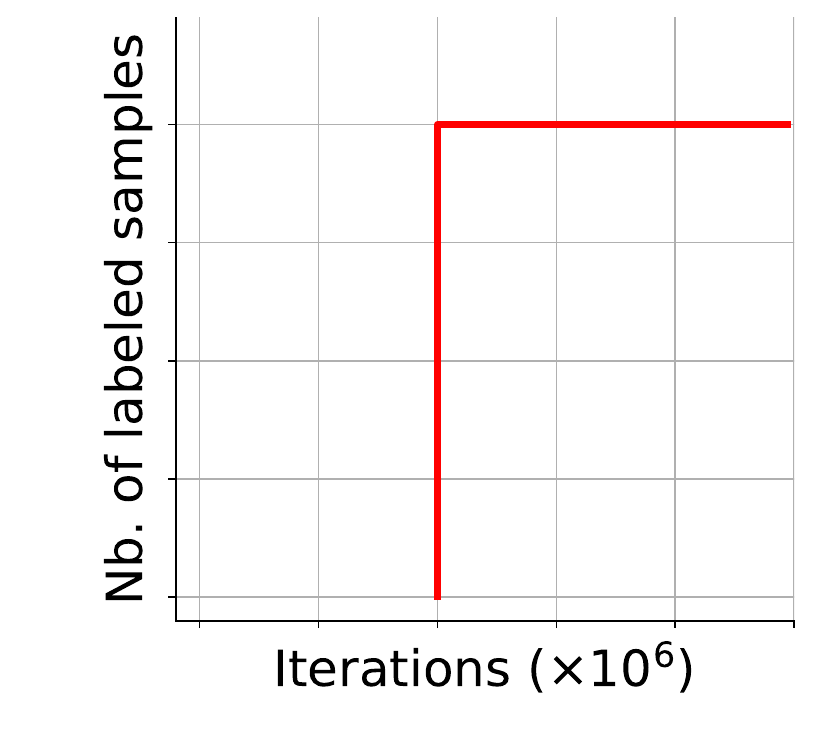}}
\subfloat[]{\includegraphics[width= 0.33\columnwidth]{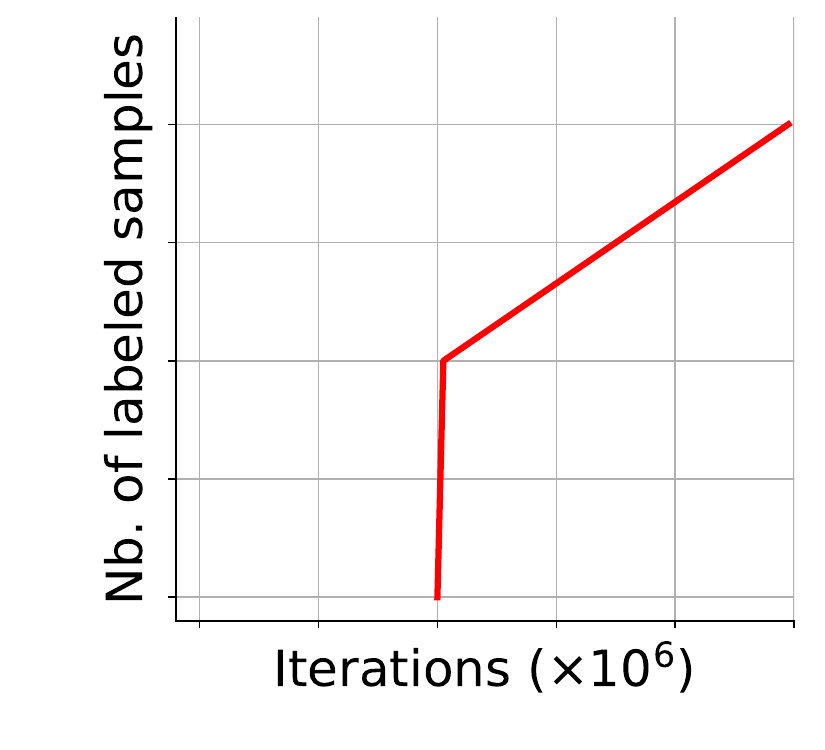}}
\caption{Illustrations of explored supervision policies. Here, (a) is the naive supervision policy, and (b) to (f) are explored policies.}
\label{fig:injection_plot}

\end{figure}

\subsubsection{Supervision Policy}
\label{sec:policy}
As mentioned earlier, the supervision policy aims at better utilizing the labelled data $X_L^*$ in the semi-supervised training by using a supervision policy $P$. We explore two broad categories for this component: (\textit{a}) incremental supervision policy, and (\textit{b}) curriculum supervision policy. We expect a gradual label injection strategy to bring two advantages to semi-supervised training. First, the injection of new data could act as a natural regularizer and prevent the model from overfitting on currently available data. Second, changing the prediction distribution at a given interval could reduce the confirmation bias problem (predicting wrong pseudo-labels with high confidence) \cite{arazo2020pseudo} of semi-supervised training.

\noindent \textit{Incremental supervision policy.}
In this approach, starting with a small portion of labelled data from $X_L^*$, we gradually increase the number of labelled samples up to $n$ over a certain number of iterations. Depending on the increasing strategy, we explore 3 major variants of this policy: (i) linear increment policy, (ii) step increment policy, and (iii) late initialized policy. In the `linear increment policy', the labelled samples are injected at a linear rate over a pre-defined period of iterations. Let $n_0$ be the number of labelled samples to start the training with, and $n$ be the total number of labelled samples. Also, let $e_0$ and $e_f$ be the iterations at which incremental injection starts and ends, and $e$ be the total number of iterations. With the linear incremental supervision policy, the total number of samples at step $i$ can be defined as:
\begin{equation}
    n_i=n_0 + \left \lfloor \frac{n-n_0}{e_f-e_0} \right \rfloor \times i.
\end{equation}
Figure \ref{fig:injection_plot} (a) demonstrates the general case no particular policy is used for injecting the labelled samples. In (b) and (c), we depict two variants of our linear increment policy, wherein (b) the labelled data are injected over the full span of training iterations ($e_f=e$), while (c) utilizes different values for $e_f$ and $n_0$, where $0<e_f<e$ and $n_0>0$.

On the other hand, in the `step increment policy', $m < n$ samples are injected at a number of steps. Under this policy, the total number of labelled samples at step $i$ is:
\begin{equation}
    n_i=n_0 + \left \lfloor \frac{n-n_0}{(e_f-e_0) \times m} \right \rfloor \times i \times m.
\end{equation}
Figure \ref{fig:injection_plot} (d) demonstrates an example of this policy where $m > 1$.

Finally, in the `late initialized policy', the utilization of labelled samples does not start until a pre-defined iteration. During this period (prior to the use of labels), the model is trained under unsupervised loss only. figures\ref{fig:injection_plot} (e) and (f) are examples of this policy where $e_0>0$ and $e_f=e_0$ for (f).

\noindent \textit{Curriculum supervision policy.}
As indicated earlier, we also explore a curriculum-based strategy as our supervision policy. Curriculum learning is often compared to human learning behaviour, where we start learning with an easy task and slowly learn more complex tasks. However, in machine learning settings, all the available data are generally used at the same time to train the model. Curriculum learning was popularized by \cite{bengio2009curriculum} in the context of machine learning. Here, pre-defined criteria are used to select easy samples at the beginning of the training and slowly introduce harder samples. Similar to the incremental supervision policy, the curriculum supervision policy also increases the number of labelled samples gradually. However, this policy uses a particular criterion to identify the order in which the labelled data is injected.  
Here, we utilize the pre-trained encoder to rank the samples. We define an `easy' sample as having the least entropy in the final prediction of the pre-trained model, while the hard samples are defined as those with higher entropy. All the policies depicted in Figure \ref{fig:injection_plot} are also explored with the curriculum policy.

{\renewcommand{\arraystretch}{1.1}
    \begin{table}[]
    \caption{Summary of training hyper-parameters.}
     
    \resizebox{0.5\textwidth}{!}{
    \setlength{\tabcolsep}{7pt}
    \begin{tabular}{l|cccc}
    \hline
    
    \textbf{Dataset} & \textbf{CIFAR-10} & \textbf{CIFAR-100} & \textbf{SVHN} & \textbf{STL-10} \\ \hline
    
    Model & WRN-28-2 & WRN-28-8& WRN-28-2 & WRN-37-2 \\
    Batch size & 64 & 64 & 64 & 64  \\
    LR & 0.03& 0.03& 0.03& 0.03\\
    Optimizer & SGD & SGD & SGD & SGD\\
    Weight decay & 5e-4 & 1e-3 & 5e-4 & 3e-4 \\
    EMA momentum & 0.999 & 0.999 & 0.999 & 0.999 \\

    \hline
    \end{tabular} 
    \label{tab_hparam}
     }
    
    \end{table}
}

\section{Experiments and Results}\label{sec:results}
In this section, we will first discuss the experiment setup, datasets and implementation details for our experiment. Next, we will discuss the findings from strategic sampling and supervision policy in subsequent subsections. 

\subsection{Experiment Setup}

We explore our investigation into the two proposed research questions on CIFAR-10 \cite{krizhevsky2009learning}, CIFAR-100 \cite{krizhevsky2009learning}, SVHN \cite{netzer2011reading}, and STL-10 \cite{coates2011analysis} datasets, which are the common datasets used by existing semi-supervised methods. Except for CIFAR-100 with 100 classes, the rest of the datasets have 10 classes each. We experiment with the strategic sampling and supervision policy modules on 8 well-known semi-supervised methods as backbones: PiModel \cite{pi_model}, PseudoLabel \cite{pseudo_labels}, MeanTeacher \cite{mean_teacher}, VAT \cite{vat}, MixMatch \cite{mixmatch}, ReMixMatch \cite{remixmatch}, UDA \cite{uda}, and FixMatch \cite{fixmatch}. Since both strategic sampling and supervision policy can be included with the existing semi-supervised models without any change, we follow all the training details and hyper-parameter settings from the original methods. For a fair comparison, the labelled and unlabelled set is kept the same across all semi-supervised methods where applicable. All the results presented in this section are run 3 times with different random initialization. The reported accuracy is average over these 3 runs.  

\subsection{Datasets}
The experiments in this paper are conducted on 4 popular datasets that are commonly used with semi-supervised learning: CIFAR-10, CIFAR-100, SVHN, and STL-10. 
\textbf{CIFAR-10} contains 60,000 samples of 10 classes with 50,000 training and 10,000 test samples. The samples have an input resolution of $32\times32$. \textbf{CIFAR-100} contains 100 classes with the same input resolution as CIFAR-10. This dataset also has 60K samples with 50K and 10K split for training and testing. \textbf{Street View House Number (SVHN)} is a dataset of digits with 10 classes that also have an input resolution of $32\times 32$. This dataset contains 73,257 samples in the training set and 26,032 samples in the testing set. Additionally, this dataset contains 531,131 samples without labels. Finally, \textbf{STL-10} is a dataset with a resolution of $96\times96$. This dataset contains 5000 training samples and 8000 test samples. It also contains an unlabelled set with 100,000 samples.

{\renewcommand{\arraystretch}{1.1}
    \begin{table*}[]
    \caption{Accuracy of strategic sampling for different clustering methods with `Inter.' and `Final' embeddings.}
     
    \setlength{\tabcolsep}{7pt}
    \begin{center}
    \begin{tabular}{l|cc|cc |cc|cc}
    \hline
    & \multicolumn{4}{c|}{\textbf{CIFAR-10}} &\multicolumn{4}{c}{\textbf{CIFAR-100}}\\
    \hline
    
     &\multicolumn{2}{c|}{\textbf{Imbalanced Split}} & \multicolumn{2}{c|}{\textbf{Balanced Split}}  &\multicolumn{2}{c|}{\textbf{Imbalanced Split}}           &\multicolumn{2}{c}{\textbf{Balanced Split}}\\
     
    \hline
    \textbf{Clustering Method }& \textbf{Inter.} & \textbf{Final} & \textbf{Inter.} & \textbf{Final} & \textbf{Inter.} & \textbf{Final} & \textbf{Inter.} & \textbf{Final}  \\
    \hline
    
    Random Sample  &   23.23$\pm1.3$ & 23.23$\pm1.3$ & 25.05$\pm1.1$ & 25.05$\pm1.1$& 11.01$\pm0.6$ & 11.01$\pm0.6$ & 11.42$\pm0.6$ & 11.42$\pm0.6$ \\ 
    \hline
    
    K-means  &  24.25$\pm1.3$ &  26.00$\pm0.8$ & 26.76$\pm0.6$  &  26.01$\pm0.6$ &\textbf{ 12.50$\pm0.2$} & 11.78$\pm0.5$ & 13.75$\pm0.8$& 13.02$\pm0.6$\\
    
    K-means++ & 23.15$\pm0.3$ & 25.69$\pm1.5$ &  26.77$\pm0.3$ & 25.83$\pm0.8$ & 12.11$\pm0.5$ & 12.13$\pm1.0$ &\textbf{ 13.95$\pm0.8$} & 13.55$\pm0.5$ \\
    
    Bisecting K-means & 21.00$\pm1.0$ & 26.17$\pm1.3$ & 24.97$\pm0.9$&  27.35$\pm0.8$  & 9.64$\pm0.6$ &  10.89$\pm0.4$  & 13.29$\pm0.4$ &  13.49$\pm0.5$\\
    
    Bisecting K-means++ & 20.83$\pm1.1$  & \textbf{26.91$\pm3.4$} & 25.39$\pm1.4$& \textbf{27.80$\pm0.8$} & 10.07$\pm0.7$ &10.55$\pm0.4$ &  13.54$\pm0.4$ & 13.64$\pm0.4$\\

    \hline
    
    & \multicolumn{4}{c|}{\textbf{SVHN}} &\multicolumn{4}{c}{\textbf{STL-10}}\\
    \hline
     &\multicolumn{2}{c|}{\textbf{Imbalanced Split}} & \multicolumn{2}{c|}{\textbf{Balanced Split}}  &\multicolumn{2}{c|}{\textbf{Imbalanced Split}}           &\multicolumn{2}{c}{\textbf{Balanced Split}}\\
     
    \hline
 & \textbf{Inter.} & \textbf{Final} & \textbf{Inter.} & \textbf{Final} & \textbf{Inter.} & \textbf{Final} & \textbf{Inter.} & \textbf{Final}  \\
    \hline
    Random Sample  &   14.35$\pm1.3$ &14.35$\pm1.3$ & 13.76$\pm1.5$ &13.76$\pm1.5$ & 24.41$\pm2.3$ &24.41$\pm2.3$ & 24.48$\pm1.6$ &24.48$\pm1.6$  \\ 
    \hline
    
    K-means & \textbf{19.04$\pm0.7$ }&   16.34$\pm1.5$ &   13.65$\pm2.0$ &   13.91$\pm2.0$ &   24.88$\pm1.7$ &   25.82$\pm2.2$ &   28.16$\pm1.1$ &   26.64$\pm1.5$  \\
    K-means++ & 18.89$\pm0.4$ &   15.98$\pm1.7$ &   \textbf{14.18$\pm2.0$} &   14.07$\pm1.3$ &   \textbf{25.32$\pm0.7$} &   \textbf{26.13$\pm2.4$ }&   28.04$\pm1.1$ &   26.35$\pm0.3$  \\
    Bisecting K-means & 15.17$\pm0.9$ &   17.88$\pm0.9$ &   12.1$\pm1.0$ &   12.22$\pm1.0$ &   23.08$\pm0.3$ &   25.02$\pm0.8$ &   \textbf{29.27$\pm2.5$ }&   25.9$\pm0.9$  \\
    Bisecting K-means++ & 16.52$\pm2.4$ &   18.31$\pm1.4$ &   11.76$\pm1.9$ &   11.85$\pm1.8$ &   21.24$\pm0.5$ &   22.39$\pm1.0$ &   28.57$\pm1.1$ &   \textbf{27.53$\pm0.6$}  \\

    \hline 
    \end{tabular}
    \label{tab_clustering_methods}
    \end{center}
    \end{table*}
}

\subsection{Implementation details}
This section describes the hyper-parameters and implementation details for all the semi-supervised methods. Following the existing semi-supervised literature, we use wide ResNet (WRN) as the encoder. More specifically, WRN-28-2 is used for CIFAR-10 and SVHN, WRN-28-8 for CIFAR-100, and WRN-37-2 for STL-10. All the models are trained using an SGD optimizer with a learning rate of 0.03 and a momentum of 0.9. The weight decay regularizer is used with a value of 1e-4 for CIFAR-100, 3e-4 for STL-10, and 5e-4 for the rest of the datasets. A batch size of 64 is used for the labelled dataset. The batch size of the unlabelled data is the same as the original SSL implementations. For example, in FixMatch, the unlabelled batch size is 7 times that of the labelled batch size. For the models that used EMA, the EMA momentum is set to 0.999. For all semi-supervised models, the unsupervised loss weight is set to 1. For pseudo-label and FixMatch, a confidence threshold value of 0.95 is used. 
This work is implemented using PyTorch, and the experiments are run on 4 Nvidia V100 GPUs. A summary of all training hyper-parameters is presented in Table \ref{tab_hparam}.

\subsection{Results of strategic sampling}
The existing literature on semi-supervised learning has reported the results with an equal number of labelled samples used per class. However, our sampling strategy, by default, selects the most diverse set of samples over the unlabelled set, which does not guarantee a balanced set of labelled samples. To thoroughly study the impact of sampling, we also modified the sampling strategy to select a balanced set of samples by applying the clustering technique to the class. Following, we present detailed results on some of the key components for strategic sampling, including the clustering algorithm and pre-trained encoder. Next, we present the results of applying strategic sampling on different semi-supervised methods.

\subsubsection{Effect of clustering algorithms}\label{sec:res:clustering}
In Table \ref{tab_clustering_methods}, we present the results on different choices for clustering using both intermediate embeddings (`Inter.') and final embeddings (`Final') for clustering. We present these results for 4 labelled samples per class for balanced, and 40, 400, 40, and 40 labelled samples for imbalanced splits of CIFAR-10, CIFAR-100, SVHN and STL-10. The observation from Table \ref{tab_clustering_methods} can be summarized by the following two observations:

\noindent (a) \textit{Final embeddings vs. intermediate embedding performance.}  For CIFAR-10 (a simpler dataset with a small number of classes), clustering with the final embedding selects more strategic sets for both imbalanced and balanced splits. On the imbalanced split, the sampling with intermediate embeddings improves the random sampling by 1.02\%, while the final embedding shows an improvement of 3.68\%. On the balanced split, the improvements are 1.72\% and 2.03\%, respectively. For the CIFAR-100 dataset (a complex dataset with more classes), using the intermediate helps with strategic sampling for both imbalanced and balanced splits. Here, with the imbalanced split, using the intermediate embeddings shows an improvement of 1.49\% over random sampling compared to the 1.12\% improvement with the final embedding. On the balanced split, using the intermediate embeddings yields a 2.53\% improvement compared to 2.22\% with the final embeddings. Similarly, for SVHN and STL-10, the intermediate embeddings show relatively better performance for all settings except STL-10 on the balanced split. Overall, SVHN shows 4.69\% and 0.42\% improvements for the imbalanced and balanced splits, while for STL-10, the improvements are 1.72\% and 4.79\%. 

\noindent (b) \textit{K-means vs. Bisecting K-means clustering.} The Bisecting K-means performs well on CIFAR-10 and STL-10, while simpler K-means clustering works better for CIFAR-100 and SVHN. For CIFAR-10, the highest accuracy is achieved by Bisecting K-means++, which is 26.91\% and 27.80\% on the imbalanced and balanced splits, respectively. On CIFAR-100, K-means and K-means++ on the imbalanced and balanced splits give the highest accuracy of 12.50\% and 13.95\%, respectively. Similarly, for SVHN, the highest accuracies of 19.04\% and 14.18\% are observed for K-means and K-means++. For STL-10, Bisecting K-means shows the best results for the balanced split. Even though the choice of clustering seems to impact the performance, we observe that using almost any clustering method is better than random sampling. 

\begin{figure}
    \centering 
    \includegraphics[width=1\linewidth]{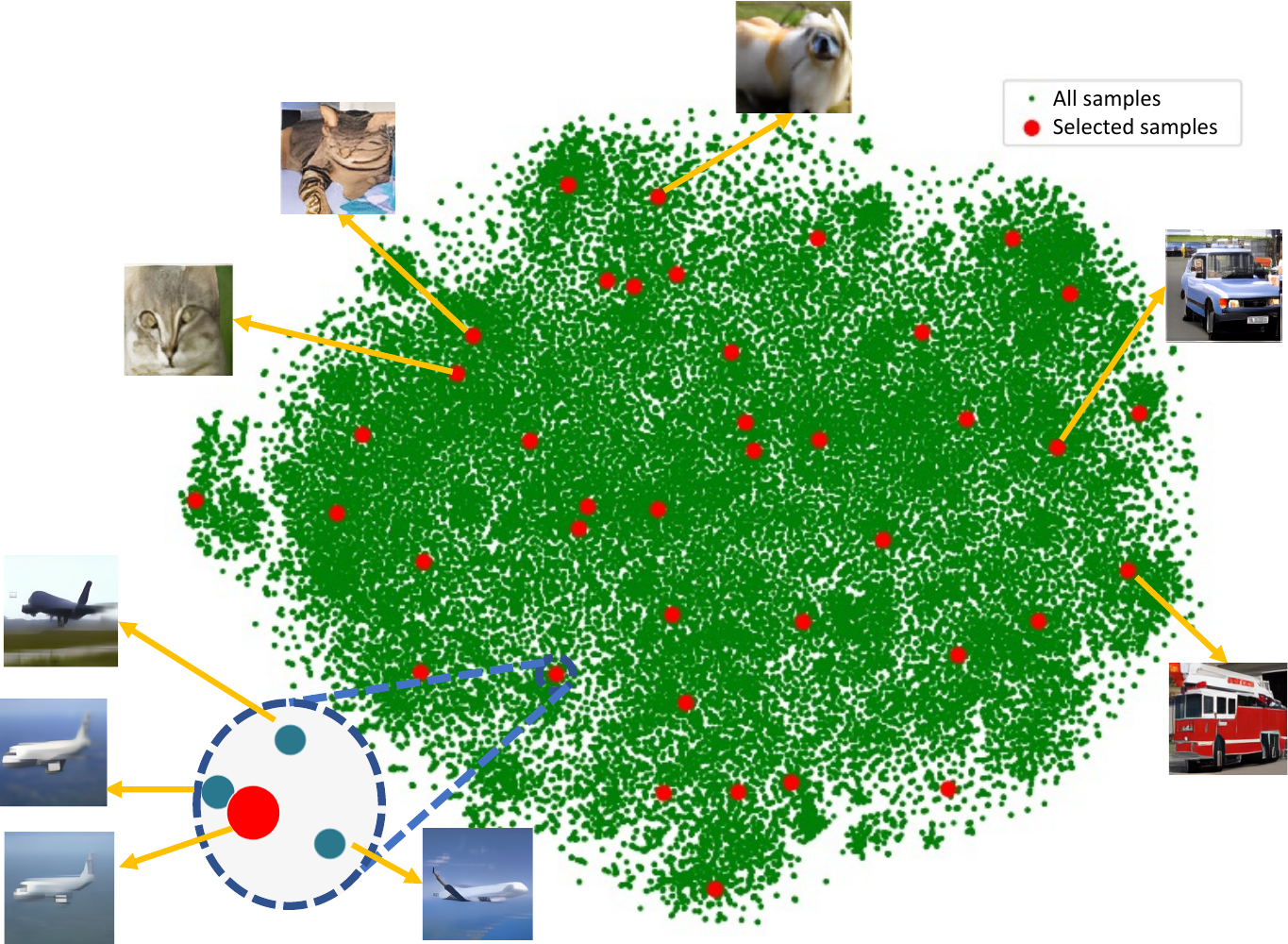}
    \caption{\textcolor{black}{A t-SNE visualization of the unlabelled and selected samples shows that the selected samples are distributed across the entire embedding space. Samples that are in close proximity to the centroids exhibit strikingly similar appearances to the centroid and are likely to contain redundant information.}}
    \label{fig:tsne}
\end{figure}

\textcolor{black}{A t-SNE visualization of the unlabelled dataset along with the selected samples, in Figure \ref{fig:tsne}, indicates that the selected samples are distributed across the entire embedding space, suggesting the diversity of the selected samples throughout the dataset. It also indicates that selected samples close in this space exhibit semantic similarities (e.g., cat and dog), while those distant from each other are semantically dissimilar (e.g., dog and truck). Samples that are in close proximity to the centroids exhibit strikingly similar appearances to the centroid and are likely to contain redundant information.}

\begin{figure}
    \centering 
    \includegraphics[width=0.7\linewidth]{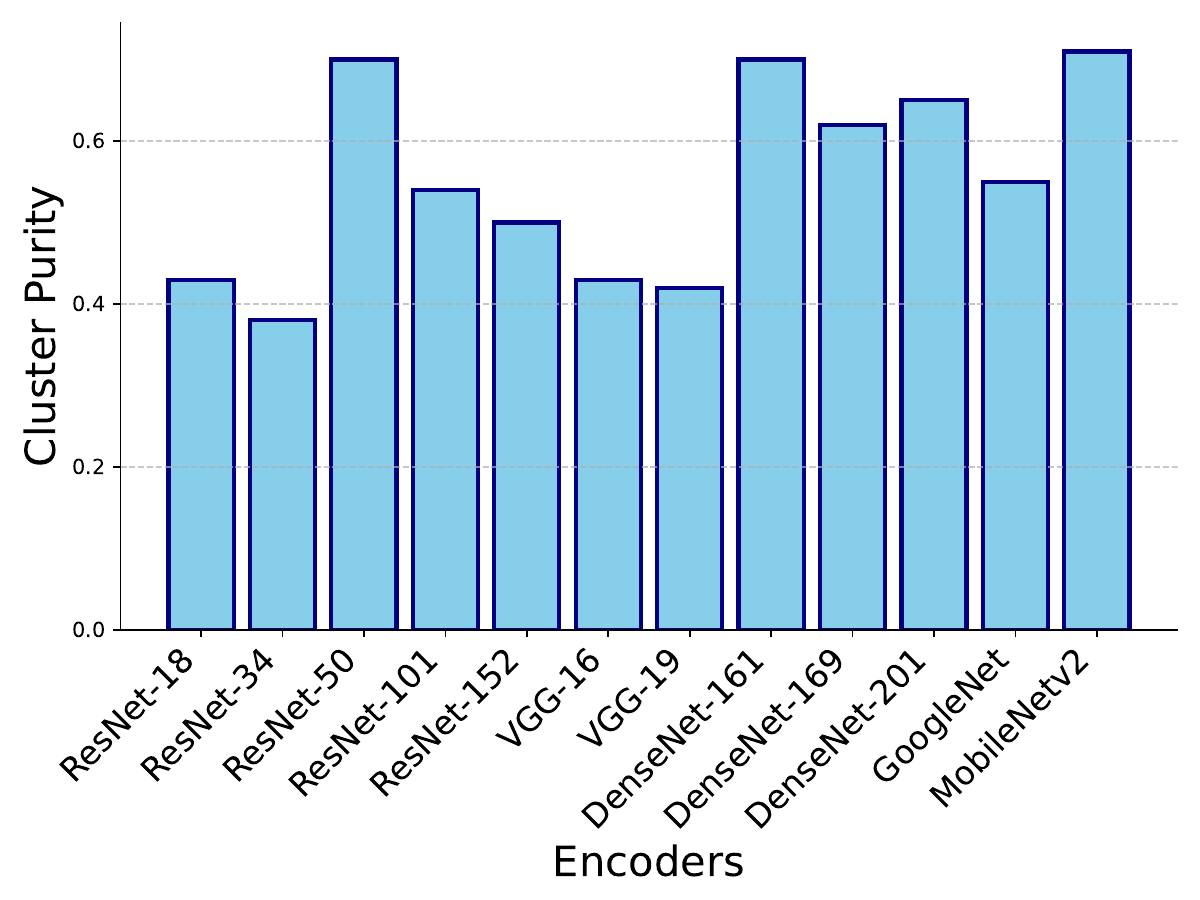}
    \caption{\textcolor{black}{Cluster purity for extracted features of different pre-trained encoders on CIFAR-10 dataset.}}
    \label{fig:purity}
\end{figure}

{\renewcommand{\arraystretch}{1.1}
    \begin{table*}[t]
     
    \caption{Accuracy of strategic sampling with different pre-trained encoders.}
    
    \begin{center}
    \begin{tabular}{l|cc |cc  | cc | cc  }
    \hline
    & \multicolumn{2}{c|}{\textbf{CIFAR-10}} &\multicolumn{2}{c|}{\textbf{CIFAR-100}} &  \multicolumn{2}{c|}{\textbf{SVHN}}  & \multicolumn{2}{c}{\textbf{STL-10}}\\
    \hline
    
     \textbf{Encoder} &\multicolumn{1}{c}{\textbf{Imbalanced}} & \multicolumn{1}{c|}{\textbf{Balanced}}  &\multicolumn{1}{c}{\textbf{Imbalanced}}           &\multicolumn{1}{c|}{\textbf{Balanced}}  &\multicolumn{1}{c}{\textbf{Imbalanced}}           &\multicolumn{1}{c|}{\textbf{Balanced}}  &\multicolumn{1}{c}{\textbf{Imbalanced}}           &\multicolumn{1}{c}{\textbf{Balanced}}\\
     
    \hline
    Rand. Sample  &  23.23$\pm1.3$ & 25.05$\pm1.1$ & 11.01$\pm0.6$ & 11.42$\pm0.6$ & 14.35$\pm1.3$ & 13.76$\pm1.5$ & 24.41$\pm2.3$ & 24.48$\pm1.6$ \\ 
    \hline
    ResNet-18 & 24.12$\pm0.3$ & 27.25$\pm0.5$ & \textbf{13.12$\pm0.5$} & 13.09$\pm0.6$& 16.48$\pm1.8$ & 	 13.01$\pm0.4$ & 	 \textbf{25.28$\pm1.6$} & 	 27.28$\pm0.3$\\
    ResNet-34 & 24.78$\pm1.7$ & 26.22$\pm0.7$ & 12.55$\pm0.3$ & \textbf{14.30$\pm0.3$}  & \textbf{18.59$\pm0.1$} & 	 12.79$\pm0.6$ & 	 24.98$\pm1.9$ & 	 27.03$\pm0.7$\\
    ResNet-50  &   \textbf{26.00$\pm0.8$}   &  26.01$\pm0.6$  &  11.78$\pm0.5$ &  13.02$\pm0.6$ & 18.59$\pm1.0$ & 	 12.78$\pm0.2$ & 	 24.21$\pm1.1$ & 	 25.64$\pm0.8$\\
    
    ResNet-101 & 22.69$\pm1.0$ & 26.43$\pm0.5$ & 12.34$\pm0.3$ & 13.54$\pm0.2$ & 14.58$\pm0.6$ & 	 11.59$\pm0.6$ & 	 24.01$\pm1.6$ & 	 \textbf{28.50$\pm1.0$}\\
    ResNet-152 & 23.79$\pm0.8$ & 27.47$\pm0.6$ & 11.81$\pm0.2$ & 13.94$\pm0.2$& 13.67$\pm1.4$ & 	 13.17$\pm0.2$ & 	 23.65$\pm1.9$ & 	 24.40$\pm0.4$\\ \hline
    VGG-16 & 23.24$\pm1.1$ & 24.35$\pm2.3$ & 11.87$\pm0.7$ & 13.69$\pm0.6$ & 15.36$\pm1.3$ & 	\textbf{ 14.18$\pm2.4$} & 	 22.85$\pm0.4$ & 	 26.33$\pm0.5$\\
    VGG-19& 24.51$\pm2.1$ & 26.42$\pm2.8$ & 11.99$\pm0.3$ & 13.85$\pm0.7$& 12.38$\pm0.4$ & 	 11.19$\pm0.3$ & 	 23.93$\pm0.8$ & 	 25.36$\pm1.4$\\ \hline
    DenseNet-161& 23.77$\pm1.5$ & 24.37$\pm0.9$ & 12.08$\pm0.3$ & 13.72$\pm0.7$& 15.17$\pm1.5$ & 	 11.24$\pm0.4$ & 	 23.09$\pm1.2$ & 	 24.22$\pm0.1$\\
    DenseNet-169& 23.99$\pm1.3$ & 22.05$\pm0.9$ & 12.69$\pm0.6$ & 13.51$\pm0.3$ & 16.05$\pm0.9$ & 	 12.36$\pm1.3$ & 	 25.34$\pm1.0$ & 	 25.72$\pm0.6$\\
    DenseNet-201& 25.51$\pm0.2$ & \textbf{28.79$\pm0.8$} & 12.17$\pm0.2$ & 12.74$\pm0.2$ & 13.04$\pm0.9$ & 	 13.43$\pm0.3$ & 	 24.98$\pm0.8$ & 	 24.26$\pm0.6$\\ \hline
    GoogleNet& 20.73$\pm1.1$ & 26.88$\pm0.5$ & 11.66$\pm0.2$ & 12.65$\pm0.1$ & 14.17$\pm1.1$ & 	 11.31$\pm0.7$ & 	 23.36$\pm1.8$ & 	 24.81$\pm1.1$\\
    MobileNetv2& 23.05$\pm0.7$ & 25.03$\pm0.1$ & 11.63$\pm0.4$ & 12.24$\pm0.9$ & 17.21$\pm1.1$ & 	 13.14$\pm0.5$ & 	 23.43$\pm2.7$ & 	 27.39$\pm0.5$\\
    
    \hline
    \end{tabular}
    \label{tab_encoder_ablation}
    \end{center}
    \end{table*}
}

\subsubsection{Performance of different pre-trained encoders}
We perform extensive experiments on the effect of different pre-trained encoders for strategic sampling. Table \ref{tab_encoder_ablation} presents the results for  ResNet-18/34/50/101/152 \cite{resnet}, VGG-16/19 \cite{vgg}, DenseNet-161/169/121 \cite{densenet}, GoogleNet \cite{googlenet} and MobileNetv2 \cite{mobilenetv2}. All these experiments are done on both imbalanced and balanced splits. For fairness and simplicity, all the experiments in this table use K-means as the clustering algorithm on the final embeddings. As we observe from this table, larger encoders like ResNet-50 and DenseNet-201 show the best results for CIFAR-10, with 26.00\% and 28.79\% accuracy on the imbalanced and balanced splits, respectively. 
However, for CIFAR-100, comparatively smaller encoders like ResNet-18 and ResNet-34 generate better-labelled sets with accuracies of 13.12\% and 14.30\%, respectively. For SVHN and STL-10, smaller encoders like ResNet18 and ResNet34 show strong performance for both datasets. However, the best performance on the balanced split for SVHN and STL-10 is observed with comparatively larger VGG16 and ResNet101 encoders, respectively. For SVHN, strategic sampling shows improvements over random sampling by 4.24\% and 0.42\% for the imbalanced and balanced splits, respectively. One unexpected observation for SVHN comes from the fact that the imbalanced split performs better than the balanced split for both random sampling and strategic sampling. This indicates that depending on the complexity and distribution of the dataset, some classes might require more labelled samples than others, and a balanced split is not always optimal. This once again shows the importance of the proposed strategic sampling that samples the most diverse samples over the entire unlabelled set. For STL-10, the imbalanced and balanced splits show 0.87\% and 4.02\% improvements over random sampling. The most important observation from this table is that almost all the encoders for every setting outperform the random sampling by a notable margin. Although the choice of encoders dictates the best accuracy for different datasets and settings, using any encoder with the strategic sampling strategy can provide tangible improvements.

\textcolor{black}{
To inspect the quality of the clustering of the extracted features, we measure their \textit{cluster purity} \cite{reddy2013data}. This measure is often used to evaluate the quality of clustering by calculating the extent to which clusters contain items from a single class. Higher values of cluster purity indicate higher quality and separability of extracted features.  Figure \ref{fig:purity} shows the cluster purity for different pre-trained encoders explored in our work. We observe that the encoders exhibit relatively similar performances in terms of this metric, indicating a consistency among clustering quality based on features obtained from different encoders.}

{\renewcommand{\arraystretch}{1.1}
    \begin{table*}[]
    \setlength
    \tabcolsep{7pt}
\caption{Accuracy of strategic sampling for different numbers of labelled samples.}

    \begin{center}
    \begin{tabular}{c|cc|cc |cc|cc}
    \hline
    & \multicolumn{4}{c|}{\textbf{CIFAR-10}} &\multicolumn{4}{c}{\textbf{CIFAR-100}}\\
    \hline
    
     &\multicolumn{2}{c|}{\textbf{Imbalanced Split}} & \multicolumn{2}{c|}{\textbf{Balanced Split}}  &\multicolumn{2}{c|}{\textbf{Imbalanced Split}}           &\multicolumn{2}{c}{\textbf{Balanced Split}}\\
     
    \hline
    \textbf{Total Samples }& \textbf{RS} & \textbf{LS} &\textbf{RS} & \textbf{LS} &\textbf{RS} & \textbf{LS} &\textbf{RS} & \textbf{LS} \\
    \hline
    1$\times C$ & 15.12$\pm2.1$ & 18.21$\pm0.8$ & 16.81$\pm1.3$ & 19.78$\pm0.4$ & 4.46$\pm0.4$ & 5.74$\pm0.4$ & 5.54$\pm0.3$ & 6.35$\pm0.3$ \\ 
    2$\times C$ & 17.68$\pm0.3$ & 20.63$\pm0.4$ & 18.32$\pm0.2$ & 24.57$\pm0.2$ & 7.07$\pm0.3$ & 8.26$\pm0.2$ & 7.83$\pm0.4$ & 9.46$\pm0.0$ \\ 
    4$\times C$  &  23.23$\pm1.3$ &  26.00$\pm0.8$ & 25.05$\pm1.1$ &  26.01$\pm0.6$ & 11.01$\pm0.6$ & 11.78$\pm0.5$ & 11.42$\pm0.6$ & 13.02$\pm0.6$\\

    10$\times C$ &31.68$\pm2.5$ & 32.26$\pm2.5$ & 31.98$\pm2.3$ & 33.57$\pm2.8$ & 21.32$\pm0.9$ & 21.37$\pm0.1$ & 21.05$\pm0.4$ & 22.24$\pm0.5$\\ 
    25$\times C$ & 41.21$\pm2.0$ & 43.16$\pm1.5$ & 41.93$\pm1.3$ & 44.05$\pm0.4$ & 36.96$\pm0.4$ & 35.3$\pm0.2$ & 37.59$\pm0.5$ & 37.51$\pm0.6$ \\
    
    100$\times C$ & 61.78$\pm0.5$ & 61.94$\pm0.5$ & 62.06$\pm0.7$ & 62.36$\pm5.3$ & 61.43$\pm0.3$ & 61.47$\pm0.3$  & 61.45$\pm0.3$ & 60.99$\pm0.3$ \\ 
    \hline
    & \multicolumn{4}{c|}{\textbf{SVHN}} &\multicolumn{4}{c}{\textbf{STL-10}}\\
    \hline
     &\multicolumn{2}{c|}{\textbf{Imbalanced Split}} & \multicolumn{2}{c|}{\textbf{Balanced Split}}  &\multicolumn{2}{c|}{\textbf{Imbalanced Split}}           &\multicolumn{2}{c}{\textbf{Balanced Split}}\\
     
    \hline
 & \textbf{RS} & \textbf{LS} &\textbf{RS} & \textbf{LS} &\textbf{RS} & \textbf{LS} &\textbf{RS} & \textbf{LS} \\
    \hline

    1$\times C$ & 12.35$\pm3.5$ & 	 13.77$\pm3.1$ & 	 12.73$\pm3.2$ & 	 13.96$\pm3.0$ & 	 17.8$\pm1.7$ & 	 19.67$\pm1.5$ & 	 17.57$\pm1.5$ & 	 18.52$\pm1.5$\\
2$\times C$ & 12.35$\pm1.9$ & 	 14.42$\pm1.8$ & 	 12.69$\pm1.7$ & 	 14.9$\pm2.0$ & 	 18.3$\pm1.1$ & 	 19.42$\pm0.8$ & 	 18.3$\pm1.1$ & 	 19.2$\pm0.7$\\
4$\times C$ & 14.78$\pm1.5$ & 	 16.76$\pm1.6$ & 	 15.04$\pm1.7$ & 	 16.14$\pm2.0$ & 	 23.96$\pm0.9$ & 	 25.78$\pm0.6$ & 	 23.89$\pm0.5$ & 	 24.24$\pm0.8$\\
8$\times C$ & 20.38$\pm2.0$ & 	 21.05$\pm2.4$ & 	 19.53$\pm2.6$ & 	 19.93$\pm2.6$ & 	 28.93$\pm0.9$ & 	 29.04$\pm0.7$ & 	 29.32$\pm0.3$ & 	 31.48$\pm0.5$\\
10$\times C$ & 24.62$\pm3.6$ & 	 25.26$\pm4.0$ & 	 24.86$\pm3.5$ & 	 25.89$\pm3.5$ & 	 31.73$\pm0.3$ & 	 32.41$\pm0.2$ & 	 31.81$\pm0.4$ & 	 32.01$\pm0.3$\\
25$\times C$ & 64.41$\pm3.4$ & 	 65.3$\pm2.0$ & 	 61.42$\pm1.0$ & 	 63.07$\pm1.3$ & 	 39.59$\pm1.3$ & 	 39.9$\pm1.1$ & 	 39.64$\pm1.3$ & 	 40.1$\pm1.2$\\
100$\times C$ & 86.89$\pm0.3$ & 	 86.92$\pm0.3$ & 	 86.89$\pm0.3$ & 	 87.68$\pm0.6$ & 	 60.49$\pm0.4$ & 	 62.12$\pm28.4$ & 	 60.77$\pm0.9$ & 	 62.49$\pm0.4$\\

    \hline
    \end{tabular}
    \label{tab_num_labels}
    \end{center}
    \end{table*}
}

\begin{figure}
\centering
\subfloat[\scriptsize{Random vs. Strategic}]{\includegraphics[width= 0.4\columnwidth]{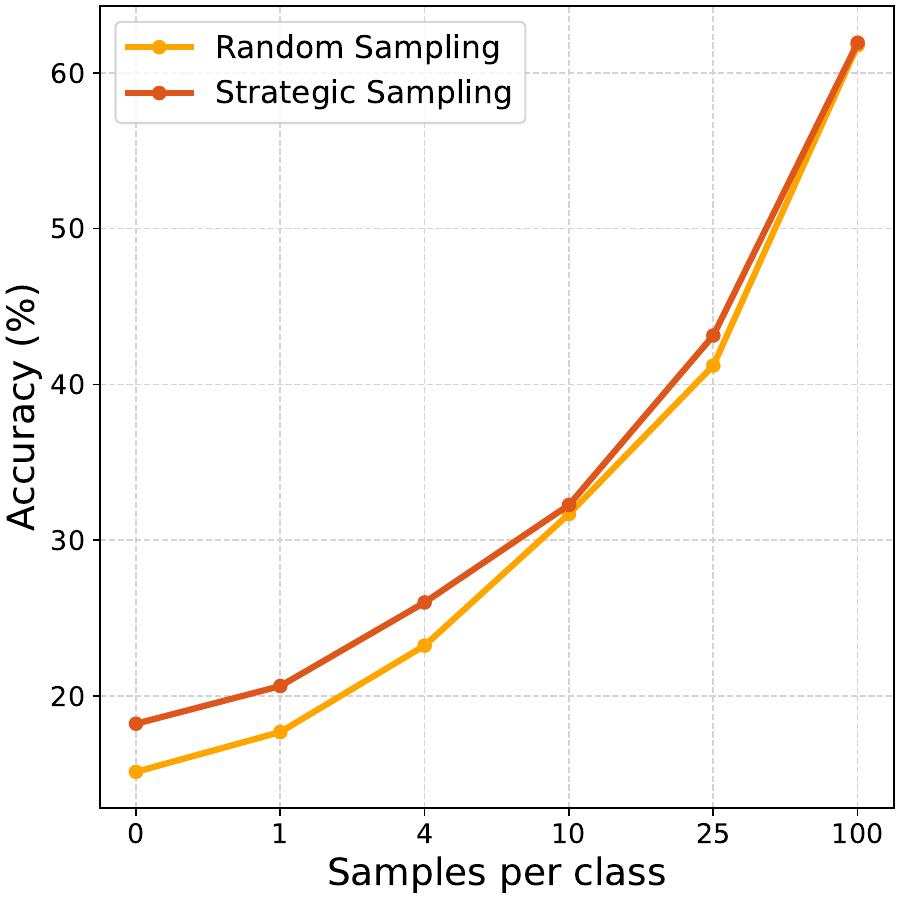}}~~~~
\subfloat[\scriptsize{Imbalanced vs. Balanced}]{\includegraphics[width= 0.4\columnwidth]{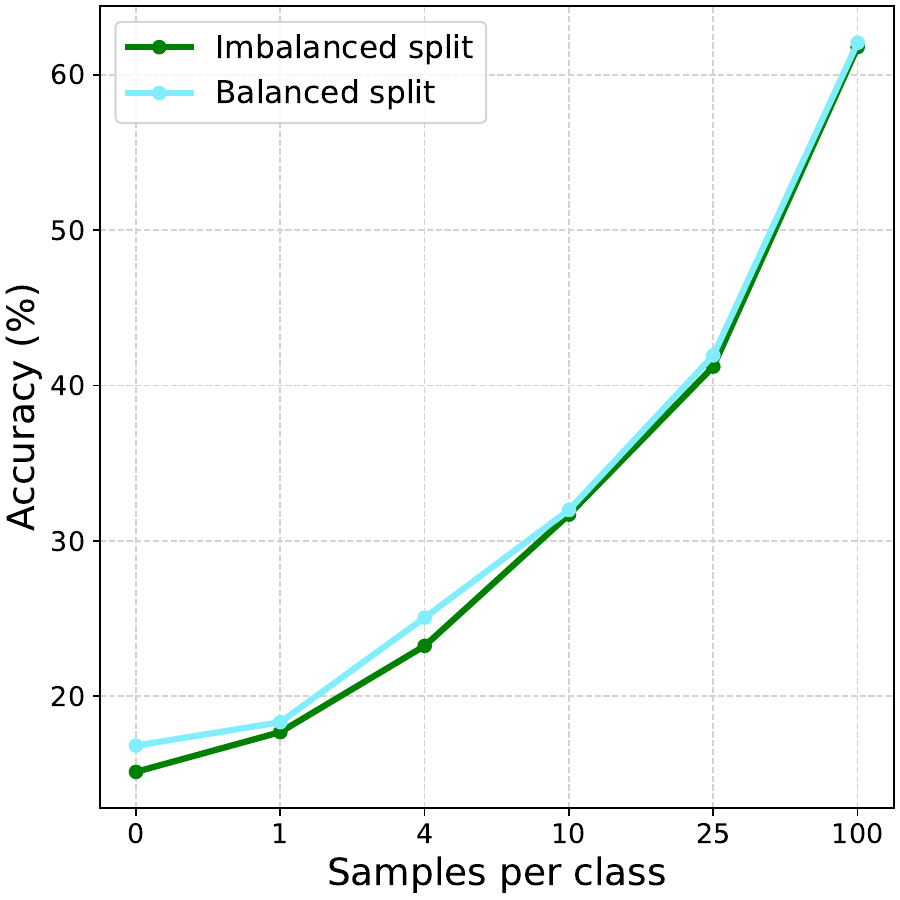}}
\caption{\textcolor{black}{The trend in accuracy for (a) random vs. strategic sampling, and (b) imbalanced vs. balanced sampling, for different numbers of samples per class.}}
\label{fig:plot_strategies}
\end{figure}

\subsubsection{strategic sampling with different labelling budgets} 
Next, in Table \ref{tab_num_labels}, we present the performance of strategic sampling when different numbers of labelled samples are used to train the model. Here, `RS' indicates random sampling, and `LS' indicates the strategic sampling strategy. All the experiments in this table are done with K-means clustering using the final embedding from ResNet-50. Note that, the optimal settings found in Table \ref{tab_clustering_methods} and \ref{tab_encoder_ablation} are not utilized here, since the purpose of this study is to understand the relative trend with the increase of data. The results from Table \ref{tab_num_labels} can be summarized in the following two observations:

{\renewcommand{\arraystretch}{1.1}   
    \begin{table*}[t!]
    \centering
     
    \caption{Accuracy of different semi-supervised methods with strategic sampling for CIFAR-10 and CIFAR-100. } 

    \setlength{\tabcolsep}{9pt}
    {
    \begin{tabular}{l|ccc|ccc}
    \hline
    & \multicolumn{3}{c|}{\textbf{CIFAR-10}}& \multicolumn{3}{c}{\textbf{CIFAR-100}} \\ 
     \hline
    \textbf{Methods} & \textbf{40 labels} & \textbf{250 labels} &\textbf{ 4000 labels} &\textbf{ 400 labels} & \textbf{2500 labels }& \textbf{10000 labels} \\ 
    \hline
    
    PiModel & 25.66$\pm1.8$ & 53.77±1.3 & 86.86$\pm0.5$ &    	 13.04$\pm0.8$ & 41.22±0.7 & 63.31$\pm0.1$  \\ 
    PiModel +\textit{LS}& 25.94$\pm1.2$  & 54.04±1.1 & 87.02$\pm0.4$ &    	 15.11$\pm0.5$ & 42.12±0.6 & 64.03$\pm0.2$  \\ 
    \hline
    
    PseudoLabel & 26.11$\pm2.2$ & 53.53±2.2 & 84.94±0.2 &   	 13.24$\pm0.5$ & 41.22±0.7 & 63.45$\pm0.2$  \\ 
    PseudoLabel +\textit{LS}& 26.78$\pm2.4$ & 53.92±2.1 & 85.92±0.2 &   	 15.71$\pm0.6$ & 42.15±0.6 & 64.23$\pm0.2$  \\ 
     \hline
     
    MeanTeacher &	29.91$\pm1.6$ &62.57±3.3 &      91.91$\pm0.2$&  	 18.89$\pm1.4$ & 54.84±1.1 &  68.21$\pm0.1$  \\ 
    MeanTeacher +\textit{LS} & 34.89$\pm1.3$  & 63.54±3.2 &      92.51$\pm0.2$&  	 17.92$\pm1.5$ & 54.80±1.0 &  68.25$\pm0.1$  \\ 
     \hline
     
    VAT &	25.34$\pm2.1$ & 58.93±1.8 & 89.47$\pm0.2$      &	 14.80$\pm1.4$ & 53.18±0.8 & 67.82$\pm0.3$ \\ 
    VAT +\textit{LS} & 22.26$\pm1.3$ & 57.83±1.5 & 89.50$\pm0.2$      &	 16.17$\pm1.8$ & 54.22±0.8 & 68.70$\pm0.2$ \\ 
    \hline
    
    MixMatch & 63.81$\pm6.5$ & 86.38±0.6& 93.31$\pm0.3$ &   	 32.41$\pm0.7$& 60.25±0.5 &72.23$\pm0.3$ \\ 
    MixMatch +\textit{LS} & 54.11$\pm1.5$ & 85.23±0.5& 93.36$\pm0.4$ &   	 34.96$\pm0.6$& 61.45±0.4 &72.95$\pm0.3$  \\ 
    \hline
    
    ReMixMatch &	90.12$\pm1.0$ & 93.74±0.1 &    95.17$\pm0.1$ &      	 57.25$\pm1.0$ & 73.96±0.4 & 79.94$\pm0.1$ \\ 
    ReMixMatch +\textit{LS} & 90.99$\pm2.4$ & 94.34±0.1 &    95.35$\pm0.1$ &  58.16$\pm1.3$ & 74.60±0.4 & 79.99$\pm0.1$  \\ 
    \hline
    
    UDA &	89.38$\pm3.8$ & 94.80±0.1 & 95.70$\pm0.1$ &       53.61$\pm1.6$ & 72.28±0.2 & 77.51$\pm0.1$ \\ 
    UDA +\textit{LS} & 89.53$\pm3.7$ & 94.92±0.1 & 95.77$\pm0.1$ &       53.64$\pm1.6$ & 72.45±0.2 & 77.72$\pm0.1$ \\ 
    \hline
    
    FixMatch &	92.53$\pm0.3$ & 95.11±0.1 & 95.71$\pm0.1$ &      	 53.58$\pm0.8$ & 71.96±0.2 & 77.80$\pm0.1$ \\ 
    FixMatch +\textit{LS} & 94.34$\pm0.9$ & 95.21$\pm$0.1 & 95.90$\pm0.1$  &53.65$\pm0.7$  & 71.98±0.2 & 77.85$\pm0.1$  \\

    \hline
    \end{tabular}

}
\label{tab_all_ssl_coresets}
\end{table*}
}

{\renewcommand{\arraystretch}{1.1}   
    \begin{table}[t!]
    \centering
    \small
    \caption{Accuracy of different semi-supervised methods with labelled set selection for SVHN and STL-10. } 

    \setlength{\tabcolsep}{2.5pt}
    {
    \begin{tabular}{l|ccc| c}
    \hline
    &  \multicolumn{3}{c|}{\textbf{SVHN}}& \multicolumn{1}{c}{\textbf{STL-10}}
     \\
    \hline
    \textbf{Methods} & 40 labels & 250 labels & 1000 labels& 1000 labels\\\hline
    
    PiModel &  32.53±0.9& 86.72±1.1 & 92.85±0.1 & 67.20±0.4\\
    PiModel +\textit{LS}&  32.93±0.8& 86.95±1.0 & 92.91±0.1 & 67.75±0.4 \\ \hline
    
    PseudoLabel &   35.42±5.6 & 84.42±0.9 & 90.61±0.3 & 67.40±0.7\\
     PseudoLabel +\textit{LS}&   35.92±3.5 & 84.72±0.8 & 91.01±0.3 & 67.99±0.6 \\ \hline
     
    MeanTeacher &	   63.90±4.0 & 96.52±0.1 &  96.72±0.1& 66.11±1.4\\
     MeanTeacher +\textit{LS} &  64.19±2.3 & 96.91±0.1 &  96.70±0.1& 66.12±1.2\\ \hline
     
    VAT &	  25.21±3.3 & 95.65±0.1 & 95.90±0.2 & 62.07±1.1\\
    VAT +\textit{LS} &   24.15±3.3 & 95.92±0.1 & 95.94±0.2 & 62.55±1.0\\ \hline
    
    MixMatch & 69.41±8.4 & 95.43±0.3 & 96.32±0.4 & 78.30±0.7 \\
    MixMatch +\textit{LS} &  69.92±5.1 & 95.78±0.3 & 96.59±0.2 & 78.35±0.6 \\ \hline
    
    ReMixMatch &  75.94±9.1 & 93.63±0.2 & 94.85±0.3  & 93.25±0.2\\
    ReMixMatch +\textit{LS} &   76.10±9.1 & 93.78±0.2 & 94.99±0.3  & 93.37±0.2 \\ \hline
    
    UDA &	 94.90±4.3 & 98.10±0.1 & 98.12±0.1 & 93.37±0.2\\
    UDA +\textit{LS} &   94.95±4.3 & 98.15$\pm$0.1  & 98.21±0.1 & 93.45±0.2\\ \hline
    
    FixMatch & 96.17±1.2 & 97.96±0.1 & 98.05±0.1 &  93.75±0.4\\
    FixMatch +\textit{LS} & 96.69±1.2 & 97.99±0.1 & 98.10±0.1 &  93.88$\pm$0.3\\

    \hline
    \end{tabular}

}
\label{tab_ssl_svhn_stl}
\end{table}
}

{\renewcommand{\arraystretch}{1.1}
    \begin{table}[]
     
    \caption{Performance on strategic sampling in very low data settings.}
    \setlength{\tabcolsep}{3pt}
    \begin{center}
    \begin{tabular}{l|l |c|c|c  }
    \hline
    
     \textbf{Method} & \textbf{Dataset} &\multicolumn{1}{c|}{\textbf{10 labels}} & \multicolumn{1}{c|}{\textbf{20 labels}}  &\multicolumn{1}{c}{\textbf{30 labels}}  \\
    
    \hline
    FixMatch & CIFAR-10 & 62.37$\pm4.9$ & 87.15$\pm1.5$ & 87.88$\pm0.9$  \\
    FixMatch + \textit{LS} &CIFAR-10 & 69.85 $\pm3.8$ & 90.17$\pm1.2$& 94.13$\pm0.8$  \\

    \hline
    FixMatch & CIFAR-100 & 21.25$\pm3.8$ &  34.58$\pm1.1$& 43.12$\pm0.9$   \\
    FixMatch + \textit{LS} & CIFAR-100 & 26.88 $\pm3.5$ & 38.39$\pm1.1$& 45.12$\pm0.9$  \\
    \hline
    
    FixMatch & SVHN & 49.57$\pm5.9$ & 89.15$\pm1.5$ & 93.44$\pm1.4$  \\
      FixMatch + \textit{LS} & SVHN & 55.14 $\pm4.7$ & 92.55$\pm1.4$& 94.13$\pm0.8$  \\
    \hline
    FixMatch & STL-10 & 32.70$\pm5.7$ & 41.25$\pm2.7$ & 57.01$\pm0.8$  \\
    FixMatch  + \textit{LS} & STL-10 & 35.77 $\pm3.4$ & 47.29$\pm1.7$& 57.98$\pm0.6$  \\
    \hline
    \end{tabular}
    
    \label{tab_coresets_small_ds}
    \end{center}
    \end{table}
}

\noindent (a) \textit{The strategic sampling strategies show higher improvements when fewer labelled samples are available.} As the number of labelled data increases, the probability also increases for random sampling to select enough good samples. For instance, the strategic sampling on CIFAR-10 with just 1 label per class observes 2.09\% and 2.97\% improvements on the imbalanced and balanced splits. However, with 100 samples per class, it is reduced to 0.16\% and 0.10\%. \textcolor{black}{ This trend is visualized in Figure \ref{fig:plot_strategies}(a).} A similar trend is seen for CIFAR-100, SVHN, and STL-10.

\noindent (b) \textit{The imbalanced and balanced splits approach similar performances when the number of labels increases.} The main difference in performance in balanced and imbalanced splits is caused by class imbalance in the sampled data points. As the number of labels per class increases, the amount of imbalance is also reduced in the randomly sampled subset (given the unlabelled dataset is also balanced). As a result, the imbalanced and balanced methods approach similar accuracies for both strategic sampling and random sampling. For example, with random sampling on CIFAR-10, the difference between imbalanced and balanced splits is reduced from 1.82\% to 0.28\% when labelled samples are increased from 4 to 100. \textcolor{black}{This trend is visualized in Figure \ref{fig:plot_strategies}(b).} Since the difference in the balanced and imbalanced splits is reduced with an increase in the number of labelled samples, for the rest of the experiments in this paper, we will follow the convention of balanced splits, which is also used in existing semi-supervised methods.

\subsubsection{Performance on different semi-supervised methods} 
We utilize all the findings from previous experiments on 8 recent and popular semi-supervised methods and summarize the results in Table \ref{tab_all_ssl_coresets} and \ref{tab_ssl_svhn_stl}. As we observe from Table \ref{tab_all_ssl_coresets}, there are good improvements in almost all the methods for both of the datasets by replacing the random sampling with strategic sampling. For 4 labelled samples per class, the highest improvement of 4.98\% for CIFAR-10 is observed with MeanTeacher. For CIFAR-100 with 4 labelled samples per class, the best improvement is 2.55\% for MixMatch. With the proposed strategic sampling, the best semi-supervised accuracy for CIFAR-10 and CIFAR-100 increased from 92.53\% to 94.34\% and from 57.25\% to 58.15\%, respectively. 

We also observe improvements for most of the semi-supervised methods when the number of labelled samples is increased. For instance, with 4000 and 10,000 labelled samples, CIFAR-10 and CIFAR-100 show the best improvements of 0.98\% and 0.88\% using PseudoLabel and VAT, respectively. However, similar to the previous observation, the improvements with strategic sampling are comparatively lower when the amount of labelled data is large. 

The results for SHVN and STL-10 are summarized in Table \ref{tab_ssl_svhn_stl}. The overall observation suggests improvements in all the semi-supervised methods for the different-sized labelled sets. For SVHN with 40 labels, the highest improvement of 0.52\% is achieved with FixMatch. On STL-10 with 1000 labels, the best improvement is 0.59\% on PseudoLabel. \textcolor{black}{Overall, Tables \ref{tab_all_ssl_coresets} and \ref{tab_ssl_svhn_stl} show a few instances where strategic sampling results in slightly lower performance compared to random sampling. These occurrences are primarily associated with SSL methods that are relatively low-performing. For instance, in the case of CIFAR-10 with 40 samples, VAT and MixMatch exhibit lower accuracies (25.34\% and 64.81\%, respectively) compared to the best-performing method, FixMatch (with 92.53\% accuracy). This suggests that VAT and MixMatch struggle to learn effective representations when provided with only 40 samples from this dataset, regardless of whether those samples are chosen randomly or strategically.}

Finally, to investigate the performance of strategic sampling in extremely low labelled data settings, in Table \ref{tab_coresets_small_ds} we show another experiment with only 1, 2 and 3 samples per class. 
For this experiment, we use FixMatch as the semi-supervised method and keep all other settings as before.
For such low data settings, strategic sampling shows very large improvements over random selection. With just 1 sample per class, the strategic sampling shows a 7.5\% improvement on FixMatch. For 2 and 3 samples per class, the improvements are 3.02\% and 6.25\%, respectively. Similarly, for CIFAR-100, SVHN, and STL-10 with only 1 label per class, the performance improvements are 5.63\%, 5.57\%, and 3.07\%, respectively.

{\renewcommand{\arraystretch}{1.1}
    \begin{table*}[t]
    \setlength
    \tabcolsep{9pt}
    \caption{Accuracy for different supervision policies. Here, `Curr.' stands for curriculum.}
    
    \begin{center}
    \begin{tabular}{l | cc | cc | cc | cc }
    \hline
     & \multicolumn{2}{c|}{\textbf{CIFAR-10}} &\multicolumn{2}{c|}{\textbf{CIFAR-100}} &\multicolumn{2}{c|}{\textbf{SVHN}} &\multicolumn{2}{c}{\textbf{STL-10}} \\ \hline
     \textbf{Policy} & No Curr. & Curr. & No Curr. & Curr. & No Curr. & Curr. & No Curr. & Curr. \\ \hline
    (a) Naive & 95.71 &95.71 & 77.80 & 77.80  & 98.10 & 98.10 & 93.88 & 93.88\\
    (b) $e_f=e$ & 95.65 & 95.68 & 77.50 & 77.60 & 98.01 & 98.11 & 93.75 & 93.85\\
    (c) $0<e_f<e$ \& $n_0>0$ &  \textbf{95.94}  &  \textbf{95.97} & \textbf{78.13} & \textbf{78.15} & \textbf{98.15} & \textbf{98.20} & \textbf{93.93} & \textbf{93.98}\\
    (d) $m>1$ & 95.73 & 95.70 & 78.11 & 78.12 & 98.02 & 98.00 & 93.74 & 93.79\\
    (e) $e_0>0$ \& $e_f=e_0$ & 95.68& 95.68& 77.77& 77.77 & 97.90 & 97.90 & 93.81 & 93.80\\
    (f) $e_0>0$ \& $e_f>e_0$ & 95.70 & 95.73& 77.74& 77.75 & 98.01 & 98.11 & 93.78 & 93.85\\ 
    
    \hline
    \end{tabular}
    \label{tab:sup_policies}
    \end{center}
    \end{table*}
}

{\renewcommand{\arraystretch}{1.1}
    \begin{table}[t]
    \caption{Acc of semi-supervised methods with the supervision policy.}
    \begin{center}
    \begin{tabular}{l| l l l l   }
    \hline
    \textbf{Methods} & \multicolumn{1}{c}{\textbf{CIFAR-10}} &\multicolumn{1}{c}{\textbf{CIFAR-100}}&\multicolumn{1}{c}{\textbf{SVHN}}&\multicolumn{1}{c}{\textbf{STL-10}}\\

    \hline 
    
    PiModel & 86.86$\pm0.5$ & 	 63.31$\pm0.1$  & 92.85±0.1 & 67.20±0.4\\
    PiModel +\textit{SP} &  88.66$\pm0.2$   & 64.53$\pm0.3$ & 93.36±0.1 & 67.70±0.3 \\ \hline   
    PseudoLabel & 84.92$\pm0.2$ & 	 63.45$\pm0.2$ & 90.61±0.3 & 67.40±0.7\\
    PseudoLabel +\textit{SP} & 85.35$\pm0.2$  & 63.94$\pm0.4$ & 90.69±0.3 & 67.52±0.6 \\ \hline
     
    MeanTeacher &	91.91$\pm0.2$ & 	 68.21$\pm0.1$ &  96.72±0.1& 66.11±1.4\\
    MeanTeacher +\textit{SP} & 92.39$\pm0.1$  & 68.56$\pm0.1$ &  96.79±0.1& 66.53±1.4 \\ \hline
    
    VAT &	89.47$\pm0.2$ & 	 67.82$\pm0.3$ & 95.90±0.2 & 62.07±1.1\\
    VAT+\textit{SP} & 89.43$\pm0.2$  & 67.79$\pm0.3$  & 95.98±0.2 & 62.76±0.9 \\ \hline 
    
    MixMatch & 93.31$\pm0.3$ & 	 72.23$\pm0.3$ & 96.32±0.4 & 78.30±0.7 \\
    MixMatch  +\textit{SP} &  93.53$\pm0.3$ & 	 72.45$\pm0.3$  & 96.72±0.4 & 78.55±0.7 \\ \hline
    
    ReMixMatch &	95.17$\pm0.1$ & 	 79.94$\pm0.1$ & 94.85±0.3  & 93.25±0.2\\
    ReMixMatch+\textit{SP}&	 95.26$\pm0.1$ & 	79.99$\pm0.1$ & 94.95±0.3  & 93.29±0.2\\ \hline
    
    UDA &	95.70$\pm0.1$ & 	 77.51$\pm0.1$ & 98.12±0.1 & 93.37±0.2\\
    UDA+\textit{SP} & 95.79$\pm0.1$ & 	 77.65$\pm0.1$ & 98.01±0.1 & 93.25±0.2\\ \hline
    
    FixMatch &	95.71$\pm0.1$ & 	 77.80$\pm0.1$ & 98.05±0.1 &  93.75±0.4\\
    FixMatch+\textit{SP} & 95.94$\pm0.1$ & 78.15$\pm0.1$ & 98.20±0.1 &  93.98±0.3  \\
    
    \hline
    \end{tabular}
    \label{tab_ssl_curriculum}
    \end{center}
    \end{table}
}

\subsection{Results of supervision policy}
In this section, we first present the results for different supervision policies mentioned in Section \ref{sec:policy} on FixMatch. Next, we show the results of different semi-supervised learners utilizing the best policy.

\subsubsection{Performance of different supervision policies}
Table \ref{tab:sup_policies} presents the results for all the policies that are explored as originally depicted in Figure \ref{fig:injection_plot}). In the table, the experiment named `Naive' refers to a no-supervision policy. The linear increment policy in (b), where the labelled data is injected over the full span of training ($e_f=e)$ shows a decrease in the accuracy for all datasets. However, the linear increment policy with $e_f<e$ and $n_0>0$ (policy (c)) shows small improvements. A small increment is also seen in the step increment policy (d) for CIFAR-10 and CIFAR-100, but not for SVHN and STL-10. Again, late initialized policies ((e) and (f)) do not show any improvements. Finally, the results with curriculum-based policies show slight improvements over most of the policies with no curriculum settings. The best accuracy is achieved for curriculum on policy (c) with an accuracy of 95.97\%, 78.13\%, 98.20\%, and 93.98\% on CIFAR-10, CIFAR-100, SVHN, and STL-10, respectively.

\subsubsection{Performance on different semi-supervised methods}
Next, we utilize the best policy from Table \ref{tab:sup_policies} on different SSL methods, and summarize the results in Table \ref{tab_ssl_curriculum}. The overall observation in the table is that while some improvements are observed on older methods like PiModel, PseudoLabel, and even MeanTeacher, for recent methods, the improvements are fairly small (around 0.2\%). Hence, we conclude that the performance of state-of-the-art semi-supervised learners does not generally improve under pre-defined supervision policies.

\begin{figure}
    \centering
    \includegraphics[width=0.4\textwidth]{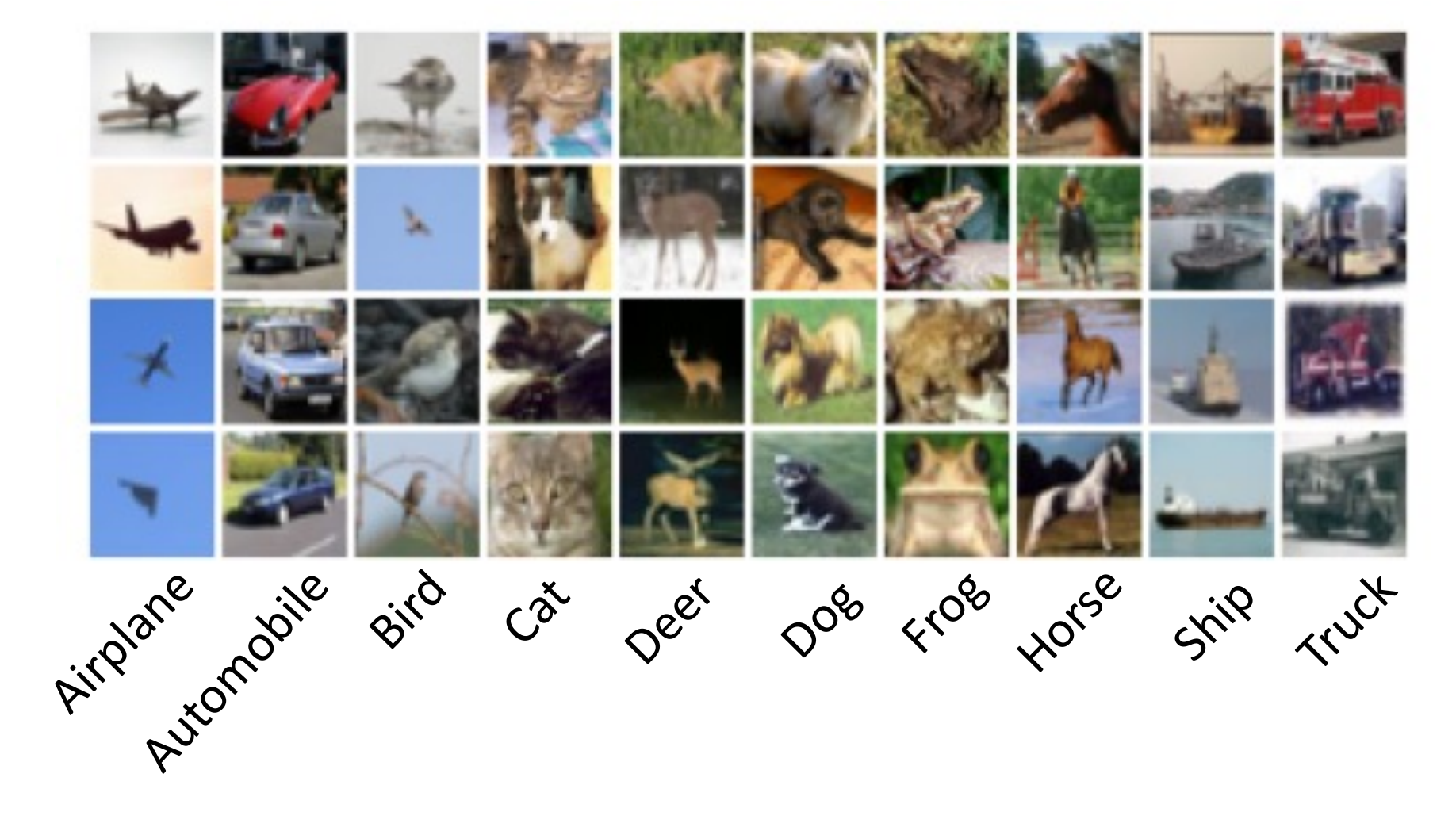}
    \caption{Examples of selected label sets with the strategic sampling from CIFAR-10.}
    \label{fig:example_label_set_cifar10}

\end{figure}

\begin{figure}
    \centering
    \includegraphics[width=0.4\textwidth]{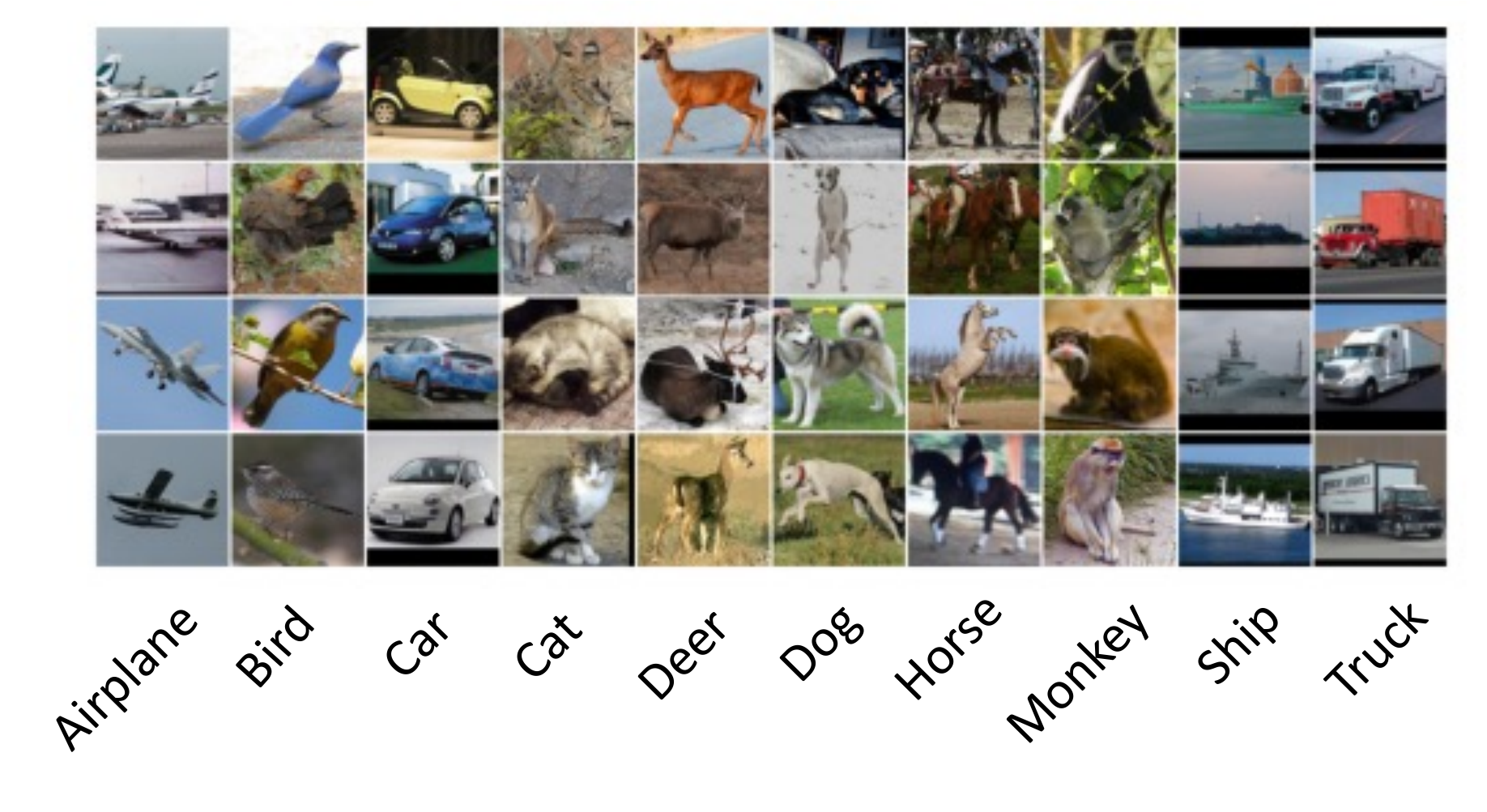}
    \caption{Examples of selected label sets with the strategic sampling from STL-10.}
    \label{fig:example_label_set_stl10}

\end{figure}

\begin{figure}
    \centering
    \includegraphics[width=0.4\textwidth]{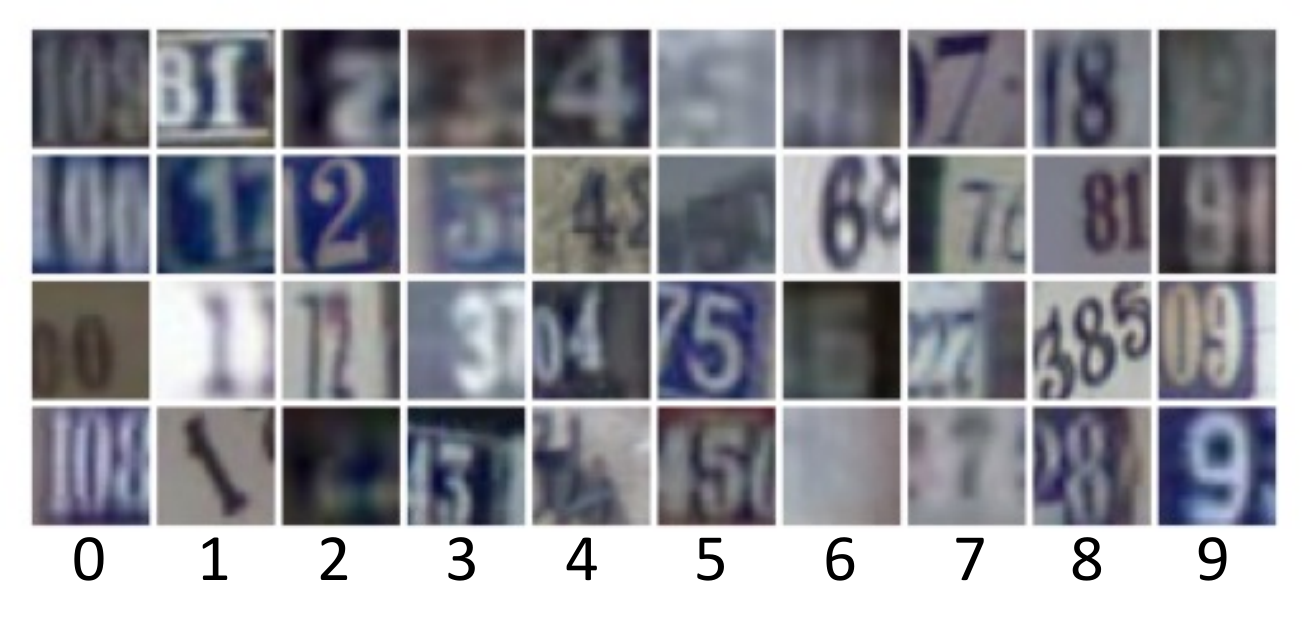}
    \caption{Examples of selected label sets with the strategic sampling from SVHN.}
    \label{fig:example_label_set_svhn}

\end{figure}

\begin{figure}
    \centering
    \includegraphics[width=0.4\textwidth]{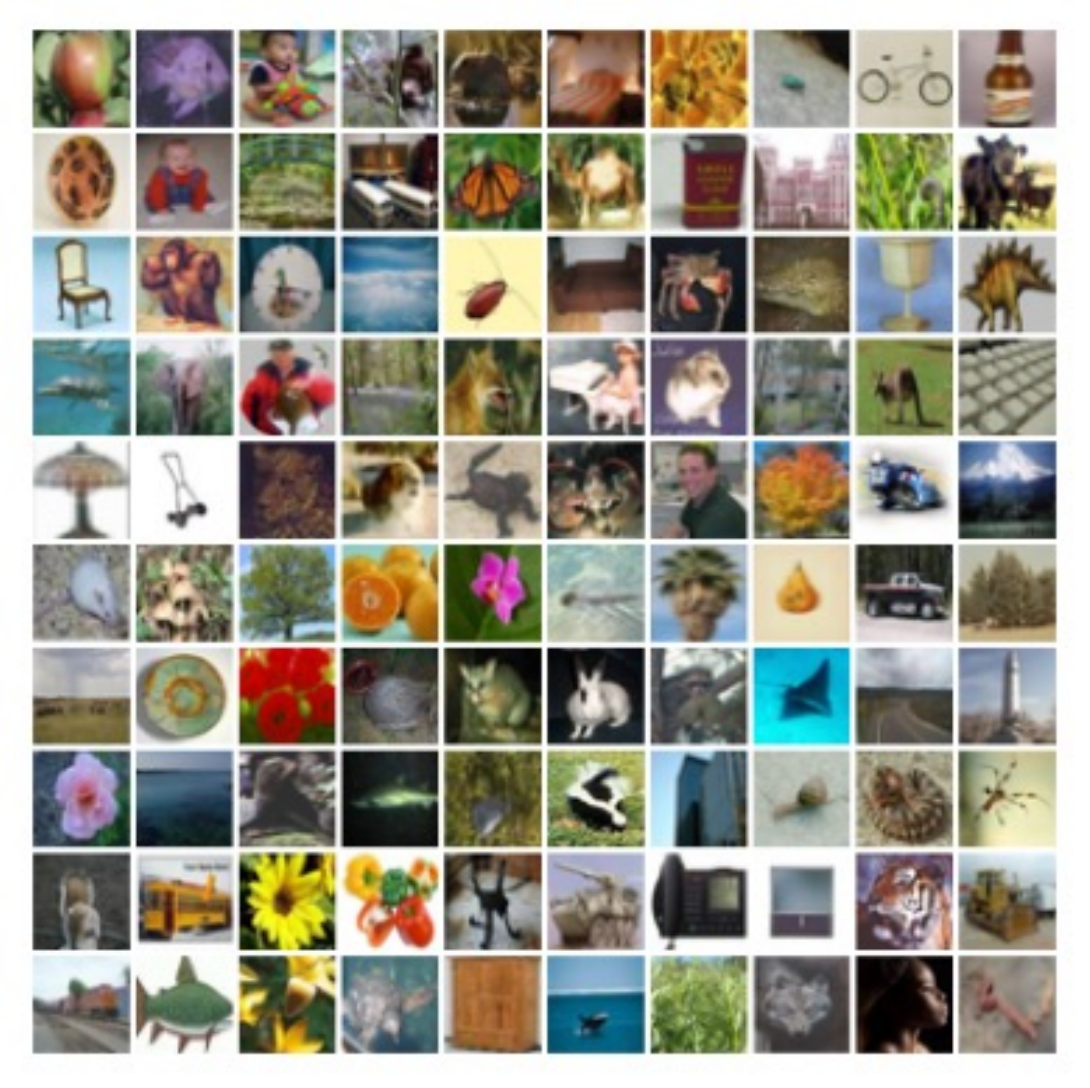}
    \caption{Examples of selected label sets with the strategic sampling from CIFAR-100.}
    \label{fig:example_label_set_cifar100}

\end{figure}

\subsection{Qualitative analysis on selected labels}
\label{app:qualitative}
Figure \ref{fig:example_label_set_cifar10} presents a qualitative analysis of the strategic sampling. Here, we show 40 samples from CIFAR-10 that are selected as the labelled set by the strategic sampling (using Bisecting K-means++ clustering on the final prediction of DenseNet201). In Figure \ref{fig:example_label_set_cifar10}, each column shows 4 samples from each of the classes. As we observe, the selected samples from each class are quite diverse in terms of visual appearance. For instance, the samples for class `airplane' include a small plane on the ground, a flying passenger plane from a side view, a flying passenger plane from a bottom view, and a flying military aircraft. Similarly, samples for the class `automobile' include samples of cars from front, back, and side views with different colours and models. A similar trend is observed for all the other classes. 

We also show similar examples for STL-10, SVHN, and CIFAR-100 in figures \ref{fig:example_label_set_stl10}, \ref{fig:example_label_set_svhn}, and \ref{fig:example_label_set_cifar100}. Similar to the observation on CIFAR-10, all the sampled images with the strategic sampling strategy show diverse and representative samples. We show only 1 image per class for CIFAR-100. 
 
\subsection{Time complexity of strategic sampling}
Strategic sampling is a one-time operation that is performed prior to training. Since our strategic sampling uses a K-means operation, which has a linear time complexity with respect to the number of input samples \cite{manning2008introduction}, the overall time complexity of our proposed strategic sampling is linear to the number of samples in the unlabeled set. As a result, our solution can be efficiently applied to large datasets without incurring significant computational overhead. As an example, whereas training of the most complex semi-supervised method in terms of computation (FixMatch) takes approximately 72 hours on a single Nvidia V100 GPU, the strategic sampling only requires 10 minutes of added time.

\section{Conclusion}
In this work, we raised two important questions regarding common practices in existing semi-supervised methods.
First, we explored whether there is a subset of unlabelled data that is more representative in comparison to random sampling, which could improve the performance of the semi-supervised learner should that subset be selected in an unsupervised fashion and labelled. Then, we explored whether there is a more efficient way of utilizing the selected labelled data rather than using all of them from the beginning of the training. Extensive studies on the first question show that careful selection of more representative and diverse samples from the unlabelled set can, in fact, improve the accuracy of almost any semi-supervised methods on any dataset. 
\textcolor{black}{This process does not introduce any computational overhead since the sampling process is a lightweight and one-time operation performed before the start of training.} To that end, with a strategic sampling strategy, we found that improvements of up to $\sim$2\% can be achieved, which could even increase to 7.5\% for very few labelled scenarios. To investigate our second question, we explored a variety of supervision policies, including pre-defined and curriculum approaches. However, we did not find any meaningful improvements over the current practice. 
\textcolor{black}{Overall, our work has a few limitations. First, the sample selection module of our solution requires a pre-trained encoder to generate the embeddings, which may not be widely available for a new task. Also, the clustering process in the strategic sampling will require a comparatively greater computational budget when dealing with larger datasets. Future work can explore strategic sampling without requiring a domain-specific pre-trained encoder. Also, more recent clustering algorithms can be explored that exhibit constant computational overheads as dataset size grows.}

\ifCLASSOPTIONcompsoc
  \section*{Acknowledgments}
\else
  \section*{Acknowledgment}
\fi

We would like to thank BMO Bank of Montreal and Mitacs for funding this research. We are also thankful to SciNet HPC Consortium for helping with the computing resources.

\ifCLASSOPTIONcaptionsoff
  \newpage
\fi

\bibliographystyle{IEEEtran}
\bibliography{ref.bib}

\begin{thebibliography}{10}
\providecommand{\url}[1]{#1}
\csname url@samestyle\endcsname
\providecommand{\newblock}{\relax}
\providecommand{\bibinfo}[2]{#2}
\providecommand{\BIBentrySTDinterwordspacing}{\spaceskip=0pt\relax}
\providecommand{\BIBentryALTinterwordstretchfactor}{4}
\providecommand{\BIBentryALTinterwordspacing}{\spaceskip=\fontdimen2\font plus
\BIBentryALTinterwordstretchfactor\fontdimen3\font minus \fontdimen4\font\relax}
\providecommand{\BIBforeignlanguage}[2]{{%
\expandafter\ifx\csname l@#1\endcsname\relax
\typeout{** WARNING: IEEEtran.bst: No hyphenation pattern has been}%
\typeout{** loaded for the language `#1'. Using the pattern for}%
\typeout{** the default language instead.}%
\else
\language=\csname l@#1\endcsname
\fi
#2}}
\providecommand{\BIBdecl}{\relax}
\BIBdecl

\bibitem{krizhevsky2012imagenet}
A.~Krizhevsky, I.~Sutskever, and G.~E. Hinton, ``Imagenet classification with deep convolutional neural networks,'' \emph{Advances in Neural Information Processing Systems}, 2012.

\bibitem{vaswani2017attention}
A.~Vaswani, N.~Shazeer, N.~Parmar, J.~Uszkoreit, L.~Jones, A.~N. Gomez, {\L}.~Kaiser, and I.~Polosukhin, ``Attention is all you need,'' \emph{Advances in Neural Information Processing Systems}, 2017.

\bibitem{fixmatch}
K.~Sohn, D.~Berthelot, N.~Carlini, Z.~Zhang, H.~Zhang, C.~A. Raffel, E.~D. Cubuk, A.~Kurakin, and C.-L. Li, ``Fixmatch: Simplifying semi-supervised learning with consistency and confidence,'' \emph{Advances in Neural Information Processing Systems}, pp. 596--608, 2020.

\bibitem{mixmatch}
D.~Berthelot, N.~Carlini, I.~Goodfellow, N.~Papernot, A.~Oliver, and C.~A. Raffel, ``Mixmatch: A holistic approach to semi-supervised learning,'' \emph{Advances in Neural Information Processing Systems}, 2019.

\bibitem{ssl_fer}
S.~Roy and A.~Etemad, ``Analysis of semi-supervised methods for facial expression recognition,'' in \emph{2022 10th International Conference on Affective Computing and Intelligent Interaction}, 2022, pp. 1--8.

\bibitem{dist_density_tails}
N.~Carlini, U.~Erlingsson, and N.~Papernot, ``Distribution density, tails, and outliers in machine learning: Metrics and applications,'' \emph{arXiv preprint arXiv:1910.13427}, 2019.

\bibitem{data_diet}
M.~Paul, S.~Ganguli, and G.~K. Dziugaite, ``Deep learning on a data diet: Finding important examples early in training,'' \emph{Advances in Neural Information Processing Systems}, pp. 20\,596--20\,607, 2021.

\bibitem{simclr}
T.~Chen, S.~Kornblith, M.~Norouzi, and G.~Hinton, ``A simple framework for contrastive learning of visual representations,'' in \emph{International Conference on Machine Learning}, 2020, pp. 1597--1607.

\bibitem{remixmatch}
D.~Berthelot, N.~Carlini, E.~D. Cubuk, A.~Kurakin, K.~Sohn, H.~Zhang, and C.~Raffel, ``Remixmatch: Semi-supervised learning with distribution alignment and augmentation anchoring,'' \emph{arXiv preprint arXiv:1911.09785}, 2019.

\bibitem{pi_model}
M.~Sajjadi, M.~Javanmardi, and T.~Tasdizen, ``Regularization with stochastic transformations and perturbations for deep semi-supervised learning,'' \emph{Advances in Neural Information Processing Systems}, 2016.

\bibitem{vat}
T.~Miyato, S.-i. Maeda, M.~Koyama, and S.~Ishii, ``Virtual adversarial training: a regularization method for supervised and semi-supervised learning,'' \emph{IEEE Transactions on Pattern Analysis and Machine Intelligence}, pp. 1979--1993, 2018.

\bibitem{mean_teacher}
A.~Tarvainen and H.~Valpola, ``Mean teachers are better role models: Weight-averaged consistency targets improve semi-supervised deep learning results,'' \emph{Advances in Neural Information Processing Systems}, 2017.

\bibitem{grandvalet2004semi}
Y.~Grandvalet and Y.~Bengio, ``Semi-supervised learning by entropy minimization,'' \emph{Advances in Neural Information Processing Systems}, 2004.

\bibitem{pseudo_labels}
D.-H. Lee \emph{et~al.}, ``Pseudo-label: The simple and efficient semi-supervised learning method for deep neural networks,'' in \emph{Workshop on Challenges in Representation Learning, ICML}, 2013, p. 896.

\bibitem{laine2016temporal}
S.~Laine and T.~Aila, ``Temporal ensembling for semi-supervised learning,'' \emph{arXiv preprint arXiv:1610.02242}, 2016.

\bibitem{uda}
Q.~Xie, Z.~Dai, E.~Hovy, T.~Luong, and Q.~Le, ``Unsupervised data augmentation for consistency training,'' \emph{Advances in Neural Information Processing Systems}, pp. 6256--6268, 2020.

\bibitem{jia2016adaptive}
L.~Jia, Z.~Zhang, L.~Wang, W.~Jiang, and M.~Zhao, ``Adaptive neighborhood propagation by joint l2, 1-norm regularized sparse coding for representation and classification,'' in \emph{IEEE International Conference on Data Mining}, 2016, pp. 201--210.

\bibitem{zhang2021flexmatch}
B.~Zhang, Y.~Wang, W.~Hou, H.~Wu, J.~Wang, M.~Okumura, and T.~Shinozaki, ``Flexmatch: Boosting semi-supervised learning with curriculum pseudo labeling,'' \emph{Advances in Neural Information Processing Systems}, pp. 18\,408--18\,419, 2021.

\bibitem{chen2023softmatch}
H.~Chen, R.~Tao, Y.~Fan, Y.~Wang, J.~Wang, B.~Schiele, X.~Xie, B.~Raj, and M.~Savvides, ``Softmatch: Addressing the quantity-quality tradeoff in semi-supervised learning,'' in \emph{International Conference on Learning Representations}, 2023.

\bibitem{fullmatch}
Y.~Chen, X.~Tan, B.~Zhao, Z.~Chen, R.~Song, J.~Liang, and X.~Lu, ``Boosting semi-supervised learning by exploiting all unlabeled data,'' in \emph{IEEE/CVF Conference on Computer Vision and Pattern Recognition}, 2023, pp. 7548--7557.

\bibitem{roy2024scaling}
S.~Roy and A.~Etemad, ``Scaling up semi-supervised learning with unconstrained unlabelled data,'' in \emph{AAAI Conference on Artificial Intelligence}, 2024, pp. 14\,847--14\,856.

\bibitem{zhang2021dual}
Y.~Zhang, Z.~Zhang, Y.~Wang, Z.~Zhang, L.~Zhang, S.~Yan, and M.~Wang, ``Dual-constrained deep semi-supervised coupled factorization network with enriched prior,'' \emph{International Journal of Computer Vision}, pp. 3233--3254, 2021.

\bibitem{zhang2017robust}
Z.~Zhang, F.~Li, L.~Jia, J.~Qin, L.~Zhang, and S.~Yan, ``Robust adaptive embedded label propagation with weight learning for inductive classification,'' \emph{IEEE Transactions on Neural Networks and Learning Systems}, no.~8, pp. 3388--3403, 2017.

\bibitem{coleman2019selection}
C.~Coleman, C.~Yeh, S.~Mussmann, B.~Mirzasoleiman, P.~Bailis, P.~Liang, J.~Leskovec, and M.~Zaharia, ``Selection via proxy: Efficient data selection for deep learning,'' \emph{arXiv preprint arXiv:1906.11829}, 2019.

\bibitem{Forgetting}
M.~Toneva, A.~Sordoni, R.~T.~d. Combes, A.~Trischler, Y.~Bengio, and G.~J. Gordon, ``An empirical study of example forgetting during deep neural network learning,'' in \emph{International Conference on Learning Representations}, 2019.

\bibitem{margatina2021active}
K.~Margatina, G.~Vernikos, L.~Barrault, and N.~Aletras, ``Active learning by acquiring contrastive examples,'' \emph{arXiv preprint arXiv:2109.03764}, 2021.

\bibitem{ducoffe2018adversarial}
M.~Ducoffe and F.~Precioso, ``Adversarial active learning for deep networks: a margin based approach,'' \emph{arXiv preprint arXiv:1802.09841}, 2018.

\bibitem{zhuang2020comprehensive}
F.~Zhuang, Z.~Qi, K.~Duan, D.~Xi, Y.~Zhu, H.~Zhu, H.~Xiong, and Q.~He, ``A comprehensive survey on transfer learning,'' \emph{Proceedings of the IEEE}, pp. 43--76, 2020.

\bibitem{steinbach2000comparison}
M.~Steinbach, G.~Karypis, and V.~Kumar, ``A comparison of document clustering techniques,'' 2000.

\bibitem{arthur2006k}
D.~Arthur and S.~Vassilvitskii, ``k-means++: The advantages of careful seeding,'' Stanford, Tech. Rep., 2006.

\bibitem{savaresi2001performance}
S.~M. Savaresi and D.~L. Boley, ``On the performance of bisecting k-means and pddp,'' in \emph{International Conference on Data Mining}, 2001, pp. 1--14.

\bibitem{boutsidis2014randomized}
C.~Boutsidis, A.~Zouzias, M.~W. Mahoney, and P.~Drineas, ``Randomized dimensionality reduction for $ k $-means clustering,'' \emph{IEEE Transactions on Information Theory}, pp. 1045--1062, 2014.

\bibitem{cohen2015dimensionality}
M.~B. Cohen, S.~Elder, C.~Musco, C.~Musco, and M.~Persu, ``Dimensionality reduction for k-means clustering and low rank approximation,'' in \emph{Annual ACM Symposium on Theory of Computing}, 2015, pp. 163--172.

\bibitem{arazo2020pseudo}
E.~Arazo, D.~Ortego, P.~Albert, N.~E. O’Connor, and K.~McGuinness, ``Pseudo-labeling and confirmation bias in deep semi-supervised learning,'' in \emph{International Joint Conference on Neural Networks}, 2020, pp. 1--8.

\bibitem{bengio2009curriculum}
Y.~Bengio, J.~Louradour, R.~Collobert, and J.~Weston, ``Curriculum learning,'' in \emph{Proceedings of the 26th annual International Conference on Machine Learning}, 2009, pp. 41--48.

\bibitem{krizhevsky2009learning}
A.~Krizhevsky, G.~Hinton \emph{et~al.}, ``Learning multiple layers of features from tiny images,'' 2009.

\bibitem{netzer2011reading}
Y.~Netzer, T.~Wang, A.~Coates, A.~Bissacco, B.~Wu, and A.~Y. Ng, ``Reading digits in natural images with unsupervised feature learning,'' 2011.

\bibitem{coates2011analysis}
A.~Coates, A.~Ng, and H.~Lee, ``An analysis of single-layer networks in unsupervised feature learning,'' in \emph{International conference on artificial intelligence and statistics}.\hskip 1em plus 0.5em minus 0.4em\relax JMLR Workshop and Conference Proceedings, 2011, pp. 215--223.

\bibitem{resnet}
K.~He, X.~Zhang, S.~Ren, and J.~Sun, ``Deep residual learning for image recognition,'' in \emph{IEEE/CVF Conference on Computer Vision and Pattern Recognition}, 2016, pp. 770--778.

\bibitem{vgg}
K.~Simonyan and A.~Zisserman, ``Very deep convolutional networks for large-scale image recognition,'' \emph{arXiv preprint arXiv:1409.1556}, 2014.

\bibitem{densenet}
F.~Iandola, M.~Moskewicz, S.~Karayev, R.~Girshick, T.~Darrell, and K.~Keutzer, ``Densenet: Implementing efficient convnet descriptor pyramids,'' \emph{arXiv preprint arXiv:1404.1869}, 2014.

\bibitem{googlenet}
C.~Szegedy, W.~Liu, Y.~Jia, P.~Sermanet, S.~Reed, D.~Anguelov, D.~Erhan, V.~Vanhoucke, and A.~Rabinovich, ``Going deeper with convolutions,'' in \emph{IEEE/CVF Conference on Computer Vision and Pattern Recognition}, 2015, pp. 1--9.

\bibitem{mobilenetv2}
M.~Sandler, A.~Howard, M.~Zhu, A.~Zhmoginov, and L.-C. Chen, ``Mobilenetv2: Inverted residuals and linear bottlenecks,'' in \emph{IEEE/CVF Conference on Computer Vision and Pattern Recognition}, 2018, pp. 4510--4520.

\bibitem{reddy2013data}
H.~V. Reddy, P.~Agrawal, and S.~V. Raju, ``Data labeling method based on cluster purity using relative rough entropy for categorical data clustering,'' in \emph{IEEE International Conference on Advances in Computing, Communications and Informatics}, 2013, pp. 500--506.

\bibitem{manning2008introduction}
C.~D. Manning, \emph{Introduction to information retrieval}.\hskip 1em plus 0.5em minus 0.4em\relax Syngress Publishing,, 2008.

\end{thebibliography}

\end{document}